\documentclass[11pt,a4paper]{article}
\usepackage{technicalreport}
\title{EvoGen-Harness: Learning Where and How to Evolve Image-Generation Harnesses}
\reportshorttitle{EvoGen-Harness}
\reporttype{Technical Report}
\reportversion{Version 1.0}       
\reportdate{September 2026}      
\reportauthors{  Jiabin Luo\textsuperscript{1}\quad
  Yinan Liu\textsuperscript{2}\quad
  Chunlei Meng\textsuperscript{3}\quad Yufei Guo\textsuperscript{1,*} }
\reportaffiliations{   \textsuperscript{1}Peking University \textsuperscript{2}Beijing University of Technology \textsuperscript{3}Fudan University} 

\reportcorrespondence{}
\reportproject{https://king-play.github.io/EvoGen-Harness/}
\reportcode{https://github.com/King-play/EvoGen-Harness}
\reportdata{}
\reportlogo{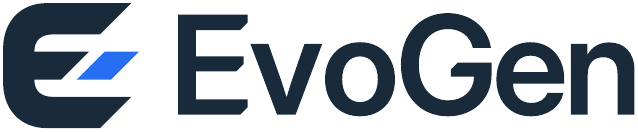}

\reportdisplaytitle{%
  \fontsize{20}{24}\selectfont
  \textsc{EvoGen-Harness}: Learning Where and How to Evolve
  Image-Generation Harnesses%
}
\hypersetup{
  pdftitle={EvoGen-Harness: Learning Where and How to Evolve Image-Generation Harnesses},
  pdfauthor={}, 
  pdfsubject={Technical report},
  pdfkeywords={text-to-image generation, harness evolution, attribution, TRACE}
}

\usepackage{amsmath,amsfonts,bm}

\def\eqref#1{equation~\ref{#1}}

\def\1{\bm{1}}

\DeclareMathAlphabet{\mathsfit}{\encodingdefault}{\sfdefault}{m}{sl}
\SetMathAlphabet{\mathsfit}{bold}{\encodingdefault}{\sfdefault}{bx}{n}

\usepackage{float}
\usepackage[most]{tcolorbox}
\tcbuselibrary{breakable}
\usepackage{booktabs,multirow,colortbl}
\usepackage{algorithm}
\usepackage{algpseudocode}
\usepackage{arydshln,tabularx,array}
\usepackage{placeins}
\newcolumntype{Y}{>{\centering\arraybackslash}X}
\newcolumntype{L}{>{\raggedright\arraybackslash}X}
\newcolumntype{P}[1]{>{\raggedright\arraybackslash}p{#1}}

\renewcommand{\footnoterule}{%
  \kern -3pt
  \hrule width 0.25\columnwidth height 0.4pt
  \kern 2.6pt
}

\begin{document}
\maketitle
\begingroup
  \renewcommand{\thefootnote}{\fnsymbol{footnote}}
  \footnotetext[1]{Corresponding Author.}
\endgroup

\begin{abstract}
Modern text-to-image (T2I) systems can be improved without modifying generator
parameters by adapting the external system around frozen generators.
However, existing approaches typically optimize a predefined dimension, such
as prompts, routing, or workflows, restricting the space in which generation
failures can be corrected.
Allowing multiple generator-external responsibilities to evolve provides a
broader adaptation space, but introduces a new challenge: visual feedback
reveals \emph{what failed}, but not \emph{where} persistent evolution should
occur or \emph{how} this space should be explored efficiently.
We introduce \textsc{EvoGen-Harness}, a generator-agnostic framework for
multi-responsibility image-generation harness evolution, together with
\textsc{Trace} (\textbf{Tra}jectory-Relative Attribution and
\textbf{C}oordinated \textbf{E}volution).
\textsc{Trace} aggregates evidence across stochastic executions, uses failure
attribution as a search prior to focus candidate updates, and progressively
re-attributes residual failures to coordinate evolution across
responsibilities, while \textsc{No-Patch} and held-out validation prevent
unnecessary or harmful updates.
Across GenEval2, T2I-CompBench++, and WISE, \textsc{EvoGen-Harness} improves
over the strongest evaluated baselines by \textbf{+0.2633},
\textbf{+0.0720}, and \textbf{+0.0752}, respectively, while achieving
\textbf{87.9--91.4\%} attribution recall, \textbf{94.8\%}
\textsc{No-Patch} accuracy, and only \textbf{1.9\%} regression.
These results demonstrate that attribution-guided multi-responsibility
evolution can substantially enhance frozen T2I systems beyond
single-dimension adaptation.

\end{abstract}
\vspace{8pt}
\begin{minipage}{\textwidth}
\captionsetup{hypcap=false}
\centering

    \centering
    \includegraphics[width=\linewidth]{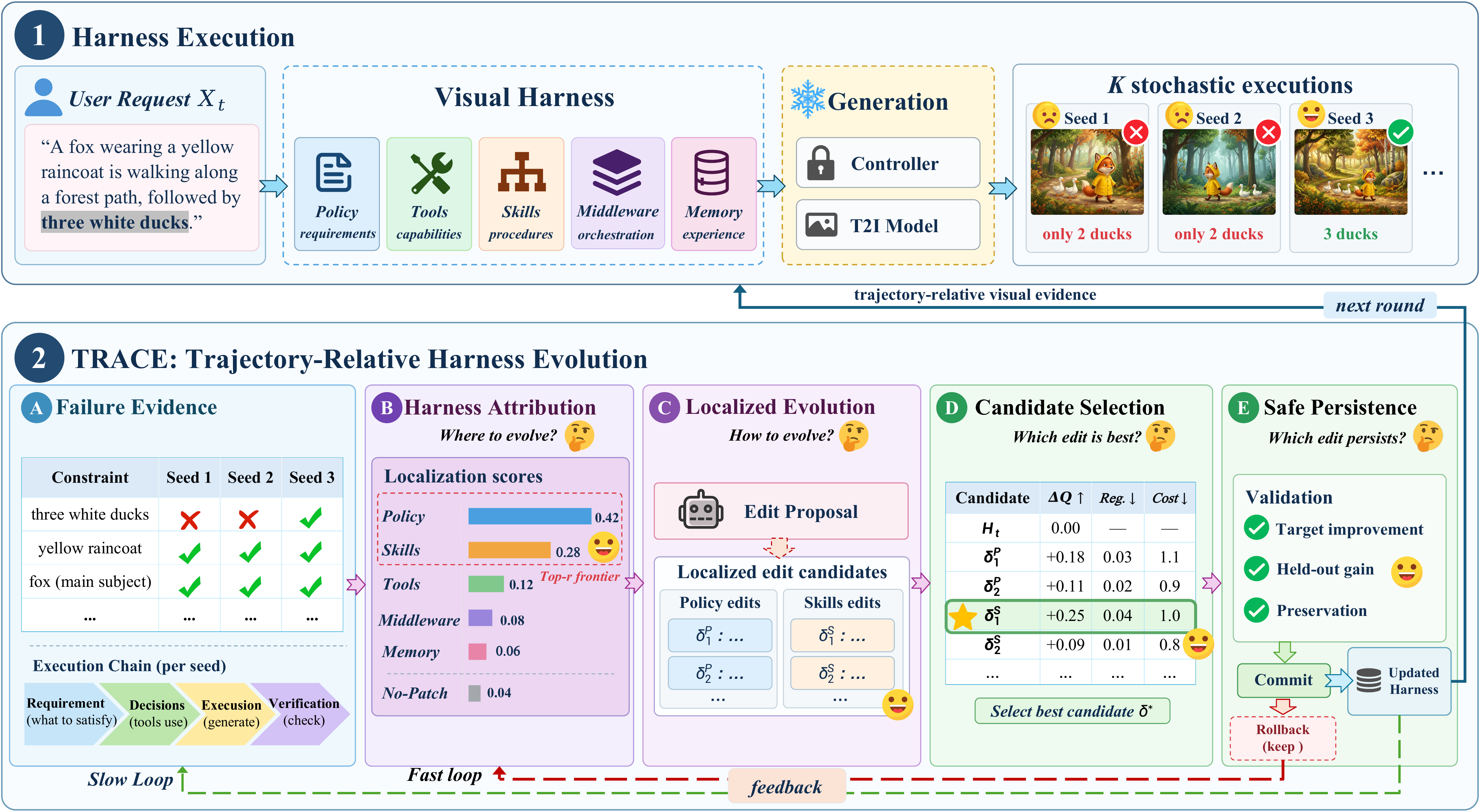}
    \captionof{figure}{
    \textbf{Overview of \textsc{EvoGen-Harness}.}
    The image generator remains frozen while its persistent
    generator-external harness evolves.
    \textsc{Trace} aggregates trajectory-relative visual evidence,
    localizes which harness responsibilities should be explored,
    proposes responsibility-conditioned edits with a fixed LLM,
    and retains only updates that pass matched evaluation and
    held-out validation.
    }
    \label{fig:overview}

\end{minipage}
\clearpage

\section{Introduction}
\label{sec:intro}

Text-to-image (T2I) generation has advanced rapidly, yet even strong diffusion
and flow-based models still struggle with compositional requirements such as
object counting, attribute binding, spatial relations, and multi-object
interactions
\citep{rombach2022ldm,esser2024sd3,ghosh2023geneval,
huang2025t2icompbenchpp,cho2024dsg}.
Recent studies show that these failures can be mitigated without modifying the
generator itself.
RePrompt and VisualPrompter improve prompt construction through reasoning and
visual feedback
\citep{wu2026reprompt,wu2026visualprompter};
DiffAgent and GenArtist coordinate generators and external tools
\citep{zhao2024diffagent,wang2024genartist};
and OctoT2I learns tool knowledge for multi-round generator routing
\citep{jiang2026octot2i}.
More generally, harness engineering suggests that persistent model-external
components can themselves become objects of optimization
\citep{lin2026ahe}.
Together, these advances establish generator-external adaptation as a promising
way to improve frozen T2I systems.

However, existing approaches largely optimize a predefined adaptation
dimension, or a small fixed subset of them.
Prompt-based methods rewrite textual inputs, routing methods reconsider
generators, while workflow- and skill-based methods modify predefined
procedures.
Such specialization can be effective, but also restricts the space in which a
generation failure can be corrected.
For example, repeated prompt refinement cannot directly repair inaccurate tool
knowledge or a flawed reusable procedure, while generator switching cannot
correct misleading long-term experience.
This motivates a broader view: instead of committing to one adaptation
dimension in advance, an image-generation agent should be able to improve
across multiple persistent responsibilities.

\begin{figure}[!b]
    \centering
    \includegraphics[width=\textwidth]{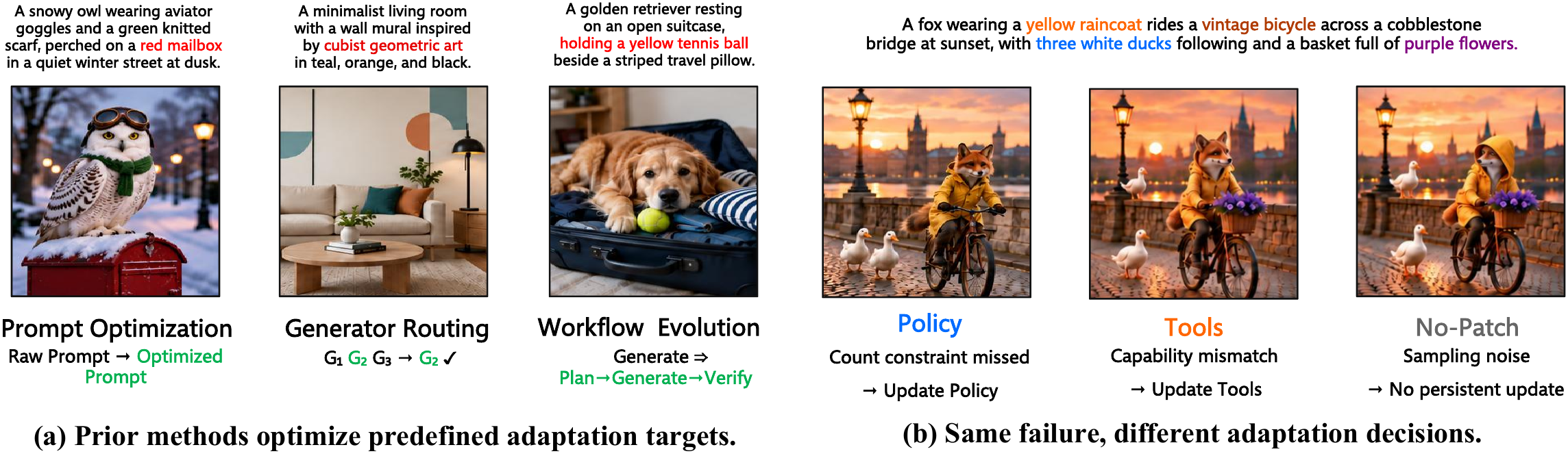}
    \caption{
    \textbf{Motivation.}
    Prior image-generation methods typically optimize a predefined adaptation
    dimension, restricting the available correction space.
    Allowing multiple harness responsibilities to evolve provides a broader
    adaptation space, but introduces a new attribution problem: the same visual
    failure may require different persistent changes, or no update at all.
    }
    \label{fig:motivation}
\end{figure}

Broadening the adaptation space, however, introduces a new challenge:
\emph{where should evolution occur, and how should this larger space be
explored efficiently?}
As illustrated in Fig.~\ref{fig:motivation}, consider a request for a fox on a
bicycle followed by \emph{three white ducks}, while only two ducks are
generated.
The same visible failure may arise because the count requirement is represented
incorrectly, the system holds inaccurate capability knowledge, a reusable
procedure or run-time decision is inadequate, previous experience is
misleading, or simply because stochastic generation produced an unlucky
sample.
Thus, visual feedback reveals \emph{what failed}, but not necessarily
\emph{what should change}.
Trying all possible harness modifications would be both expensive and
unreliable.
Multi-responsibility evolution therefore requires both
\emph{automatic attribution} and \emph{guided search}, rather than brute-force
trial and error.

To address this challenge, we introduce \textsc{EvoGen-Harness}, a
generator-agnostic framework for multi-responsibility evolution around frozen image
generators.
We operationalize persistent generator-external decisions through five
functional responsibilities:
\textbf{Policy} for visual requirements,
\textbf{Tools} for available capabilities,
\textbf{Skills} for reusable procedures,
\textbf{Middleware} for run-time orchestration,
and \textbf{Memory} for cross-task experience.
These responsibilities define a practical adaptation space rather than an
exhaustive taxonomy.
To evolve this space efficiently, we introduce
\textsc{Trace} (\textbf{Tra}jectory-Relative Attribution and
\textbf{C}oordinated \textbf{E}volution).
\textsc{Trace} aggregates constraint-level evidence across stochastic executions,
attributes failures to plausible responsibilities, and uses this attribution
as a search prior to prune unlikely edits.
Candidate updates are evaluated relative to the current harness under matched
generation conditions; after each accepted update, residual failures are
re-attributed, allowing evolution to progressively move across
responsibilities.
\textsc{No-Patch} avoids unnecessary updates, while held-out validation
prevents harmful changes from persisting.

We evaluate \textsc{EvoGen-Harness} on three complementary T2I benchmarks,
where it consistently improves over the strongest evaluated baselines, and further analyze attribution, preservation, efficiency, and cross-generator generalization.

Our contributions are summarized as follows:
\begin{itemize}
    \item We propose \textsc{EvoGen-Harness}, a
    \textbf{generator-agnostic framework for multi-responsibility
    image-generation harness evolution}.
    By exposing multiple persistent generator-external responsibilities for
    adaptation, it moves beyond optimization confined to a predefined
    dimension while keeping the underlying image generator frozen.

    \item We introduce \textbf{\textsc{Trace}}, a trajectory-relative
    attribution and coordinated evolution mechanism.
    \textsc{Trace} uses LLM-based harness attribution as a search prior,
prunes the heterogeneous edit space, evaluates candidate updates under matched
stochastic conditions, and progressively re-attributes residual failures to
enable efficient cross-responsibility evolution.

    \item Extensive experiments show that \textsc{EvoGen-Harness} improves
    over the strongest evaluated baselines by
    \textbf{+0.2633}, \textbf{+0.0720}, and \textbf{+0.0752}
    on GenEval2, T2I-CompBench++, and WISE, respectively.
    It further achieves \textbf{87.9--91.4\%} attribution recall,
    \textbf{94.8\%} \textsc{No-Patch} accuracy, and only
    \textbf{1.9\%} regression, while generalizing across frozen T2I backbones.
\end{itemize}
\section{Related Work}
\label{sec:related_work}

\subsection{Prompt Optimization and Visual Feedback}

Prompt optimization improves T2I generation while keeping the generator fixed.
Early methods learn or search for model-preferred prompts
\citep{hao2023promptist,cao2023beautifulprompt},
while recent approaches increasingly exploit visual feedback.
RePrompt learns reasoning-augmented prompt refinement from image-level rewards
\citep{wu2026reprompt}, VisualPrompter performs targeted refinement from missing
visual concepts \citep{wu2026visualprompter}, and IterComp learns
composition-aware feedback for iterative improvement
\citep{zhang2025itercomp}.
Iterative self-correction is also explored in prior work
\citep{wu2024selfcorrecting,yang2024idea2img}.
These methods demonstrate the value of visual feedback, but mainly optimize a
predefined target---the prompt or current generation.
In contrast, \textsc{EvoGen-Harness} determines which persistent
generator-external responsibility should evolve.

\subsection{Agentic and Self-Evolving Image Generation}

Agentic T2I systems broaden adaptation beyond prompts through planning, tools,
routing, and reusable experience.
LayoutGPT and LLM-grounded Diffusion use LLMs for explicit visual or layout
planning
\citep{feng2023layoutgpt,lian2023llmgrounded},
while RPG, CompAgent, Ranni, Marmot, MCCD, and M3 improve complex generation
through regional planning, decomposition, or multi-agent correction
\citep{yang2024mastering,wang2024divide,feng2024ranni,
sun2025marmot,li2025mccd,yang2026m3}.
Other agentic systems perform generator selection, tool use, and stateful routing
\citep{zhao2024diffagent,wang2024genartist,jiang2026octot2i},
while recent methods evolve workflows, skills, or visual experience
\citep{chen2026genrouter,li2026comfyclaw,chen2026genevolve}.
Despite increasing adaptivity, these approaches generally predefine
\emph{what} is improved.
Our work instead determines both \emph{where} and \emph{how} persistent
evolution should proceed.

\subsection{Agent Harness Evolution}

Recent work extends self-improvement to the model-external harness.
Agentic Harness Engineering evolves heterogeneous harness components through
observability-driven feedback \citep{lin2026ahe}, while Self-Harness improves
operating harnesses through weakness mining, modification, and regression
validation \citep{zhang2026selfharness}.
JIT-Agent further synthesizes and self-evolves task-adaptive harnesses on demand
\citep{zhang2026jitagent}.
These studies establish harness evolution as a general agentic paradigm.
\textsc{EvoGen-Harness} focuses on stochastic image generation, where
\textsc{TRACE} aggregates trajectory-relative visual evidence, guides search
over multiple responsibilities, and progressively re-attributes residual
failures to coordinate their evolution.
\section{Method}
\label{sec:method}

We present \textsc{EvoGen-Harness}, a framework that enables frozen
image-generation systems to evolve through a persistent external harness.
Instead of modifying generator parameters, \textsc{Trace}
(\textbf{Tra}jectory-Relative Attribution and
\textbf{C}oordinated \textbf{E}volution) identifies which harness component
should evolve from visual execution evidence and iteratively refines it
through feedback-driven updates.

\subsection{Multi-Responsibility Visual Harness}
\label{sec:harness}

Motivated by recent advances in self-evolving image generation, we
identify five persistent responsibilities that define the evolution space:
\textbf{Policy}, \textbf{Tools}, \textbf{Skills}, \textbf{Middleware},
and \textbf{Memory}.

At evolution round $t$, we represent the persistent generator-external state as
\begin{equation}
    \mathcal{H}_t
    =
    \bigl(
        \Pi_t,\,
        \mathcal{T}_t,\,
        \mathcal{S}_t,\,
        \mathcal{W}_t,\,
        \mathcal{M}_t
    \bigr),
    \label{eq:harness}
\end{equation}
where \textbf{Policy} $\Pi$ specifies visual requirements,
\textbf{Tools} $\mathcal{T}$ describe available capabilities,
\textbf{Skills} $\mathcal{S}$ encode reusable procedures,
\textbf{Middleware} $\mathcal{W}$ controls run-time orchestration,
and \textbf{Memory} $\mathcal{M}$ stores cross-task experience.
We denote the editable responsibilities by
$\mathcal{R}=\{\Pi,\mathcal{T},\mathcal{S},\mathcal{W},\mathcal{M}\}$.
Together, they define what the system should achieve, what capabilities it can
use, how generation should be executed, and what experience should persist.

For request $x_t$, the current harness is executed under $K$ stochastic
generation conditions $\{\omega_k\}_{k=1}^{K}$:
\begin{equation}
    \bigl(I_t^{(k)},\tau_t^{(k)}\bigr)
    =
    \operatorname{Execute}
    \bigl(x_t;\mathcal{H}_t,\mathcal{G},\omega_k\bigr),
    \qquad
    k=1,\ldots,K ,
    \label{eq:execute}
\end{equation}
where $\mathcal{G}$ is the frozen generation system,
$I_t^{(k)}$ is the generated image, and
$\tau_t^{(k)}$ records the harness decisions and executions that produced it.
The stochastic condition $\omega_k$ captures generation randomness, such as a
random seed.

\subsection{Trajectory-Relative Harness Attribution}
\label{sec:attribution}

A single failed image is insufficient evidence for persistent evolution:
the same harness may succeed under another stochastic execution.
We therefore decompose $x_t$ into fine-grained visual constraints
$\mathcal{C}_t=\{c_{t,1},\ldots,c_{t,m}\}$ and aggregate their outcomes across
the $K$ executions.
Following structured visual evaluation~\citep{cho2024dsg}, we construct
\begin{equation}
    \mathcal{E}_t
    =
    \left\{
        \left(
            c_{t,j},
            \tau_t^{(k)}\!\downarrow c_{t,j},
            v_{t,j}^{(k)}
        \right)
    \right\}_{\substack{j=1,\ldots,m\\k=1,\ldots,K}},
    \label{eq:evidence}
\end{equation}
where
$\tau_t^{(k)}\!\downarrow c_{t,j}$
denotes the trajectory decisions relevant to constraint $c_{t,j}$, and
$v_{t,j}^{(k)}$ is its visual verification result.
Thus, $\mathcal{E}_t$ connects each visual requirement to the decisions and
outcomes that may explain its success or failure.

\paragraph{LLM-based Harness Localizer.}
Given $\mathcal{E}_t$ and the current harness, a fixed LLM serves as the
\emph{Harness Localizer}.
Using a structured localization prompt, it assigns a score
$s_{\mathrm{loc}}(z\mid\mathcal{E}_t,\mathcal{H}_t)$ to each
$z\in\mathcal{R}\cup\{\varnothing\}$,
where $\varnothing$ denotes \textsc{No-Patch}.
Rather than committing immediately to one responsibility,
\textsc{Trace} retains a compact search frontier:
\begin{equation}
    \mathcal{R}_t^{(r)}
    =
    \underset{z\in\mathcal{R}}{\operatorname{Top}\text{-}r}
    \;
    s_{\mathrm{loc}}
    \bigl(
        z\mid\mathcal{E}_t,\mathcal{H}_t
    \bigr),
    \label{eq:frontier}
\end{equation}
where $r$ controls the number of plausible responsibilities retained.
\textsc{No-Patch} is selected when the evidence is better explained by
generation stochasticity, verifier uncertainty, or insufficient support for a
persistent update.

The Harness Localizer remains fixed throughout evolution and requires no
additional training.
Controlled single-responsibility interventions are used only to evaluate
localization accuracy in Sec.~\ref{sec:attribution_exp}.
Visual verification therefore identifies \emph{what failed}, while harness
localization determines \emph{where evolution should be explored}.

\subsection{\textsc{Trace}: Attribution-Guided Progressive Evolution}
\label{sec:trace}

Starting from $\mathcal{H}^{(0)}=\mathcal{H}_t$,
\textsc{Trace} explores only the responsibilities retained by the Harness
Localizer.
For each responsibility $z$, let $\Delta_z$ contain at most $M$ localized
candidate edits:
\begin{equation}
    \Delta_{\ell}
    =
    \bigcup_{z\in\mathcal{R}_{\ell}^{(r)}} \Delta_z,
    \qquad
    \Delta_z
    =
    \{\delta_{z,1},\ldots,\delta_{z,M}\},
    \label{eq:candidates}
\end{equation}
where $\delta_{z,i}$ is the $i$-th candidate edit associated with
responsibility $z$.
Depending on the selected responsibility, an edit may revise requirement
interpretation, capability knowledge, reusable procedures, run-time
orchestration, or persistent experience.

\paragraph{Responsibility-conditioned proposal.}
A fixed LLM serves as the edit proposer.
It receives the current harness state, trajectory-relative evidence, and a
responsibility-specific edit schema, and generates $\Delta_z$ while being
restricted to responsibility $z$.
The localizer and proposer may share the same underlying LLM but use separate
prompts and structured output schemas.
Thus, localization determines \emph{where} to evolve, whereas proposal
determines \emph{how} that responsibility may be modified.

\paragraph{Trajectory-relative evaluation.}
Because image generation is stochastic, independently sampled outputs can
confound edit quality with generation variance.
Let $Q(x;\mathcal{H},\omega)$ denote the visual evaluation score obtained by
executing harness $\mathcal{H}$ on request $x$ under stochastic condition
$\omega$.
Each candidate is compared with its parent harness under the same $K$
generation conditions:
\begin{equation}
\begin{aligned}
    \Delta Q_{\ell}(\delta)
    =
    \frac{1}{K}
    \sum_{k=1}^{K}
    \Big[
        &Q\!\left(
            x_t;
            \mathcal{H}^{(\ell)}\!\oplus\delta,
            \omega_k
        \right)
        \\
        &-
        Q\!\left(
            x_t;
            \mathcal{H}^{(\ell)},
            \omega_k
        \right)
    \Big].
\end{aligned}
    \label{eq:relative-gain}
\end{equation}
Here $\oplus$ denotes applying an edit to the current harness.
Matched stochastic conditions make
$\Delta Q_{\ell}(\delta)$ reflect the effect of the edit rather than random
variation in generation.

\paragraph{Progressive evolution.}
Within the attribution-pruned search space,
\textsc{Trace} ranks candidate edits by their relative improvement while
penalizing preservation regression and additional execution cost:
\begin{equation}
\begin{aligned}
    U_{\ell}(\delta)
    &=
    \Delta Q_{\ell}(\delta)
    -
    \lambda R_{\mathrm{pres}}(\delta)
    -
    \mu C(\delta),
    \\
    \delta_{\ell}^{*}
    &=
    \arg\max_{\delta\in\Delta_{\ell}}
    U_{\ell}(\delta),
    \qquad
    \mathcal{H}^{(\ell+1)}
    =
    \mathcal{H}^{(\ell)}
    \oplus
    \delta_{\ell}^{*}.
\end{aligned}
    \label{eq:progressive-evolution}
\end{equation}
Here $R_{\mathrm{pres}}(\delta)$ measures degradation of previously successful
behavior, $C(\delta)$ measures additional execution cost, and
$\lambda,\mu\geq0$ control their relative penalties.
Attribution therefore guides \emph{where to search}, while execution feedback
determines \emph{which edit to prefer}.

After each accepted edit, the updated harness is executed again and residual
failures are re-localized.
If another persistent weakness remains, \textsc{Trace} continues from
$\mathcal{H}^{(\ell+1)}$.
The resulting evolution path
\[
    \boldsymbol{\pi}
    =
    (\delta_0^{*},\ldots,\delta_{L-1}^{*})
\]
may therefore traverse multiple responsibilities, where $L$ is the maximum
evolution depth.

Evolution stops when \textsc{No-Patch} is selected, no candidate yields
positive utility, or the evolution budget is exhausted.
The resulting path is retained only if it improves the triggering and held-out
cases while keeping preservation regression below tolerance $\epsilon$;
otherwise the original harness $\mathcal{H}_t$ is restored.

Overall, \textsc{Trace} separates four complementary decisions:
visual verification determines \emph{what failed},
the Harness Localizer determines \emph{where to evolve},
the LLM proposer determines \emph{how to evolve},
and validation determines \emph{what persists}. See Appendix~\ref{app:trace} for details.
\section{Experiments}
\label{sec:experiments}

We evaluate \textsc{EvoGen-Harness} in terms of generation quality, harness
attribution, and generalization. Beyond measuring performance across diverse
T2I benchmarks, we examine whether \textsc{EvoGen-Harness} can correctly localize
persistent failures, avoid unnecessary updates, and produce reusable improvements on held-out cases and consistent gains across frozen T2I backbones.

\begin{table}[!htbp]
\centering
\caption{
\textbf{Quantitative evaluation on the GenEval2 benchmark.}
We report atom-level Soft-TIFA AM across five visual skills and prompt-level Soft-TIFA GM as the overall score.
The best and second-best results are marked in
\textbf{\textcolor{red}{bold red}} and
\underline{\textcolor{blue}{underlined blue}}, respectively.
$\uparrow$ indicates higher is better.
}
\label{tab:geneval2}

\setlength{\tabcolsep}{6.5pt}
\renewcommand{\arraystretch}{1.08}

\resizebox{\textwidth}{!}{
\begin{tabular}{clcccccc}
\toprule
\textbf{Type}
& \textbf{Method}
& \textbf{Object}
& \textbf{Attribute}
& \textbf{Count}
& \textbf{Position}
& \textbf{Verb}
& \textbf{Overall$\uparrow$} \\
\midrule

\multirow{6}{*}{\rotatebox{90}{\textit{T2I Models}}}
& FLUX.1-dev
& 0.8754 & 0.6734 & 0.5311 & 0.3928 & 0.2150 & 0.2115\\

& Janus-Pro
& 0.9102 & 0.7213 & 0.4834 & 0.5217 & 0.2056 & 0.1878 \\

& Flow-GRPO
& 0.9175 & 0.7438 & 0.6032 & 0.5233 & 0.2541 & 0.2198 \\

& Qwen-Image
& 0.9730 & 0.8412 & \underline{\textcolor{blue}{0.6857}} & 0.6083 & 0.3752 & 0.3483 \\

& Z-Image
& \underline{\textcolor{blue}{0.9755}} & 0.7851 & 0.6232 & 0.6041 & 0.2419 & 0.3034 \\

& Nano Banana
& 0.9722 & \underline{\textcolor{blue}{0.8981}} & 0.6830 & \textbf{\textcolor{red}{0.7044}} & \textbf{\textcolor{red}{0.4532}} & \underline{\textcolor{blue}{0.4456}} \\

\midrule

\multirow{4}{*}{\rotatebox{90}{\textit{Prompt Opt.}}}
& NeuroPrompts
& 0.8433 & 0.6421 & 0.5008 & 0.3712 & 0.2045 & 0.1976 \\

& Promptist
& 0.8533 & 0.6510 & 0.5122 & 0.3819 & 0.2134 & 0.2056 \\

& BeautifulPrompt
& 0.5713 & 0.4406 & 0.3567 & 0.2523 & 0.1432 & 0.1134 \\

& VisualPrompter
& 0.9321 & 0.7223 & 0.5614 & 0.4350 & 0.2412 & 0.2552 \\

\midrule

\multirow{5}{*}{\rotatebox{90}{\textit{Agentic}}}
& Idea2Img
& 0.7114 & 0.4016 & 0.2450 & 0.1967 & 0.0145 & 0.0586\\

& GenArtist
& 0.6598 & 0.3813 & 0.2535 & 0.2516 & 0.0945 & 0.0556 \\

& ChatGen
& 0.5274 & 0.3045 & 0.2011 & 0.1569 & 0.0269 & 0.0388 \\

& OctoT2I
& 0.9429 & 0.7602 & 0.5381 & 0.5506 & 0.2099 & 0.2895 \\

\rowcolor{oursbg}
& \textbf{Ours}
& \textbf{\textcolor{red}{0.9799}}
& \textbf{\textcolor{red}{0.9469}}
& \textbf{\textcolor{red}{0.9053}}
& \underline{\textcolor{blue}{0.6418}}
& \underline{\textcolor{blue}{0.3851}}
& \textbf{\textcolor{red}{0.7089}} \\

\bottomrule
\end{tabular}
}
\end{table}
\subsection{Experimental Settings}
\label{sec:exp_settings}

\paragraph{Datasets and metrics.}
We evaluate on three complementary T2I benchmarks:
GenEval2~\citep{kamath2025geneval},
T2I-CompBench++~\citep{huang2025t2icompbenchpp},
and WISE~\citep{niu2026wise}.
We follow the official evaluation protocol of each benchmark, reporting Soft-TIFA AM for skill-wise analysis and Soft-TIFA GM for overall performance on GenEval2, the official compositional metrics for T2I-CompBench++, and WiScore for WISE.
For \textsc{Trace}-specific analyses, we additionally report attribution and
evolution statistics, whose definitions are provided in
Appendix~\ref{app:metrics}.
Efficiency is measured by average wall-clock inference time and the numbers of
generator and LLM/VLM calls per prompt.

\paragraph{Baselines.}
We compare \textsc{EvoGen-Harness} with representative methods from three
categories.
For T2I generation, we include
FLUX.1-dev~\citep{flux2024},
Janus-Pro~\citep{chen2025janus},
Flow-GRPO~\citep{liu2025flow},
Qwen-Image~\citep{wu2025qwenimage},
Z-Image~\citep{team2025zimage},
and Gemini 2.5 Flash Image (Nano Banana)~\citep{google2025nanobanana}.
For prompt optimization, we compare with
NeuroPrompts~\citep{rosenman2024neuroprompts},
Promptist~\citep{hao2023promptist},
BeautifulPrompt~\citep{cao2023beautifulprompt},
and VisualPrompter~\citep{wu2026visualprompter}.
For agentic image generation, we include
Idea2Img~\citep{yang2024idea2img},
GenArtist~\citep{wang2024genartist},
ChatGen~\citep{jia2025chatgen},
and OctoT2I~\citep{jiang2026octot2i}. Unless otherwise specified, we evaluate all baselines using their official implementations and recommended configurations.

\paragraph{Implementation details.}
We use FLUX.1-dev as the frozen image generator, GPT-4.1 for harness
localization and edit proposal, and OWLv2-base with NVILA-Lite-2B-Verifier
for visual verification.
Full settings are provided in Appendix~\ref{app:implementation}.
All experiments are conducted on 2 NVIDIA A100 (80\,GB) GPUs.

\begin{table}[t]
\centering
\caption{
\textbf{Quantitative evaluation results on T2I-CompBench++.}
The best and second-best results are marked in
\textbf{\textcolor{red}{bold red}} and
\underline{\textcolor{blue}{underlined blue}}, respectively.
$\uparrow$ indicates higher is better.
}
\label{tab:t2icompbench}

\setlength{\tabcolsep}{4.8pt}
\renewcommand{\arraystretch}{1.05}

\resizebox{\textwidth}{!}{
\begin{tabular}{lccccccccc}
\toprule
\textbf{Method}
& \textbf{Color}
& \textbf{Shape}
& \textbf{Texture}
& \textbf{2D-Spatial}
& \textbf{3D-Spatial}
& \textbf{Numeracy}
& \textbf{Non-Spatial}
& \textbf{Complex}
& \textbf{Average$\uparrow$} \\
\midrule

\multicolumn{10}{c}{\textit{T2I Models}} \\
\midrule

FLUX.1-dev
& 0.7572
& 0.5066
& 0.6300
& 0.2700
& 0.3992
& 0.6165
& 0.3065
& 0.3628
& 0.4811 \\

Janus-Pro
& 0.5145
& 0.3323
& 0.4069
& 0.1566
& 0.2753
& 0.4406
& 0.3137
& 0.3806
& 0.3526 \\

Flow-GRPO
& 0.8379
& 0.6130
& 0.7236
& \underline{\textcolor{blue}{0.5447}}
& 0.4471
& 0.6752
& 0.3195
& 0.3741
& 0.5669 \\

Qwen-Image
& 0.8392
& 0.5941
& 0.7568
& 0.4512
& 0.4631
& 0.7705
& 0.3126
& 0.3944
& 0.5727 \\

Z-Image
& 0.8413
& 0.6078
& 0.7512
& 0.3822
& 0.4412
& 0.6676
& 0.3181
& 0.4201
& 0.5537 \\

Nano Banana
& \textbf{\textcolor{red}{0.8728}}
& \underline{\textcolor{blue}{0.6420}}
& \underline{\textcolor{blue}{0.7855}}
& 0.5057
& \underline{\textcolor{blue}{0.5221}}
& \underline{\textcolor{blue}{0.7923}}
& \underline{\textcolor{blue}{0.3254}}
& \underline{\textcolor{blue}{0.4239}}
& \underline{\textcolor{blue}{0.6087}} \\

\midrule
\multicolumn{10}{c}{\textit{Prompt Optimization}} \\
\midrule

NeuroPrompts
& 0.6700
& 0.4650
& 0.5200
& 0.1850
& 0.3750
& 0.5600
& 0.2860
& 0.3150
& 0.4220 \\

Promptist
& 0.6936
& 0.4827
& 0.5437
& 0.1997
& 0.3936
& 0.5874
& 0.2912
& 0.3334
& 0.4407 \\

BeautifulPrompt
& 0.5610
& 0.3940
& 0.4530
& 0.1630
& 0.3080
& 0.4380
& 0.3040
& 0.3100
& 0.3664 \\

VisualPrompter
& 0.7700
& 0.5150
& 0.6400
& 0.2850
& 0.4150
& 0.6550
& 0.3100
& 0.3850
& 0.4969 \\

\midrule
\multicolumn{10}{c}{\textit{Agentic Methods}} \\
\midrule

Idea2Img
& 0.4897
& 0.3442
& 0.3911
& 0.1473
& 0.3407
& 0.4584
& 0.2895
& 0.2945
& 0.3444 \\

GenArtist
& 0.4786
& 0.5492
& 0.5711
& 0.4716
& 0.2895
& 0.5674
& 0.2826
& 0.3723
& 0.4478\\

ChatGen
& 0.3619
& 0.3062
& 0.2767
& 0.1073
& 0.2641
& 0.3860
& 0.2931
& 0.2644
& 0.2825 \\

OctoT2I
& 0.8410
& 0.6283
& 0.7259
& 0.5256
& 0.3872
& 0.7320
& 0.3182
& 0.3582
& 0.5646 \\

\rowcolor{oursbg}
\textbf{Ours}
& \underline{\textcolor{blue}{0.8671}}
& \textbf{\textcolor{red}{0.7120}}
& \textbf{\textcolor{red}{0.7913}}
& \textbf{\textcolor{red}{0.6524}}
& \textbf{\textcolor{red}{0.6125}}
& \textbf{\textcolor{red}{0.8344}}
& \textbf{\textcolor{red}{0.4312}}
& \textbf{\textcolor{red}{0.5451}}
& \textbf{\textcolor{red}{0.6807}} \\

\bottomrule
\end{tabular}
}
\end{table}
\begin{table}[t]
\centering
\caption{
\textbf{Quantitative evaluation on the WISE benchmark.}
We report world-knowledge consistency across six domains and the overall score.
The best and second-best results are marked in
\textbf{\textcolor{red}{bold red}} and
\underline{\textcolor{blue}{underlined blue}}, respectively.
$\uparrow$ indicates higher is better.
}
\label{tab:wise}

\footnotesize
\setlength{\tabcolsep}{5pt}

\resizebox{\linewidth}{!}{%
\begin{tabular}{lccccccc}
\toprule
\textbf{Method}
& \textbf{Culture}
& \textbf{Time}
& \textbf{Space}
& \textbf{Biology}
& \textbf{Physics}
& \textbf{Chemistry}
& \textbf{Overall$\uparrow$} \\
\midrule

Qwen-Image
& 0.6275
& 0.5250
& 0.5583
& 0.3417
& 0.4833
& 0.2500
& 0.5100 \\

Nano Banana
& \underline{\textcolor{blue}{0.7000}}
& \underline{\textcolor{blue}{0.5900}}
& \underline{\textcolor{blue}{0.6700}}
& \underline{\textcolor{blue}{0.4500}}
& \underline{\textcolor{blue}{0.5800}}
& \underline{\textcolor{blue}{0.4000}}
& \underline{\textcolor{blue}{0.6028}} \\

VisualPrompter
& 0.5800
& 0.4500
& 0.5900
& 0.2300
& 0.4300
& 0.2900
& 0.4708 \\

OctoT2I
& 0.5350
& 0.4160
& 0.5090
& 0.2620
& 0.5130
& 0.3140
& 0.4557 \\

\rowcolor{oursbg}
\textbf{Ours}
& \textbf{\textcolor{red}{0.7589}}
& \textbf{\textcolor{red}{0.6712}}
& \textbf{\textcolor{red}{0.7160}}
& \textbf{\textcolor{red}{0.5889}}
& \textbf{\textcolor{red}{0.6651}}
& \textbf{\textcolor{red}{0.4788}}
& \textbf{\textcolor{red}{0.6780}} \\

\bottomrule
\end{tabular}%
}
\end{table}

\subsection{Main Results}
\label{sec:main_results}

Table~\ref{tab:geneval2} evaluates whether \textsc{EvoGen-Harness} improves
fine-grained instruction following.
Our method achieves an overall Soft-TIFA GM score of \textbf{0.7089}, outperforming both the strongest standalone T2I model (\textbf{0.4456}) and all prompt-optimization and agentic baselines.
The gain is particularly large on counting, improving from 0.6857 to
\textbf{0.9053}.
This suggests that multi-responsibility evolution is especially effective when
generation failures involve explicit visual constraints that cannot be reliably
resolved through a single predefined adaptation dimension.

Table~\ref{tab:t2icompbench} further evaluates complex compositions with
multiple interacting requirements.
\textsc{EvoGen-Harness} achieves the best average score of \textbf{0.6807}
(\textbf{+0.0720}), leading in seven of eight dimensions and showing consistent
gains as compositional difficulty increases.

On WISE (Table~\ref{tab:wise}), our method reaches \textbf{0.6780}
(\textbf{+0.0752}) and achieves the best result across all six domains.
Together, these results show that \textsc{EvoGen-Harness} is not limited to a
single failure type, generalizing from fine-grained visual constraints to
complex compositions and world-knowledge consistency.

\begin{table}[t]
\centering
\caption{
\textbf{Generalization across frozen T2I backbones on GenEval2.}
We apply the same generator-external adaptation methods to different
frozen T2I backbones and we report Soft-TIFA AM for skill-wise analysis and Soft-TIFA GM for overall performance.
$\Delta$ denotes the absolute gain of \textsc{EvoGen-Harness} over
baseline on the corresponding backbone.
The best results within each backbone are marked in \textbf{bold}.
$\uparrow$ indicates higher is better.
}
\label{tab:backbone_generalization}

\small
\setlength{\tabcolsep}{7pt}
\renewcommand{\arraystretch}{1.05}

\begin{tabular}{lcccccc}
\toprule
Method
& Object
& Attribute
& Count
& Position
& Verb
& Overall $\uparrow$ \\
\midrule

FLUX.1-dev
& 0.8754 & 0.6734 & 0.5311 & 0.3928 & 0.2150 & 0.2115 \\
+ Promptist
& 0.8533 & 0.6510 & 0.5122 & 0.3819 & 0.2134 & 0.2056 \\
+ VisualPrompter
& 0.9321 & 0.7223 & 0.5614 & 0.4350 & 0.2412 & 0.2552 \\
+ \textbf{Ours}
& \textbf{0.9799}
& \textbf{0.9469}
& \textbf{0.9053}
& \textbf{0.6418}
& \textbf{0.3851}
& \textbf{0.7089} \\
\hdashline
\rowcolor{oursbg}
\textit{$\Delta$ over VisualPrompter}
& +0.0478 & +0.2246 & +0.3439 & +0.2068 & +0.1439 & +0.4537 \\

\midrule

Qwen-Image
& 0.9730 & 0.8412 & 0.6857 & 0.6083 & 0.3752 & 0.3483 \\
+ Promptist
& 0.9484 & 0.8132 & 0.6613 & 0.5914 & 0.3724 & 0.3386 \\
+ VisualPrompter
& 0.9853 & 0.8650 & 0.7060 & 0.6355 & 0.3961 & 0.3844 \\
+ \textbf{Ours}
& \textbf{0.9956}
& \textbf{0.9742}
& \textbf{0.9365}
& \textbf{0.7689}
& \textbf{0.5106}
& \textbf{0.7594} \\
\hdashline
\rowcolor{oursbg}
\textit{$\Delta$ over VisualPrompter}
& +0.0103 & +0.1092 & +0.2305 & +0.1334 & +0.1145 & +0.3750 \\

\midrule

Janus-Pro
& 0.9102 & 0.7213 & 0.4834 & 0.5217 & 0.2056 & 0.1878 \\
+ Promptist
& 0.8872 & 0.6973 & 0.4662 & 0.5072 & 0.2041 & 0.1826 \\
+ VisualPrompter
& 0.9511 & 0.7630 & 0.5168 & 0.5549 & 0.2321 & 0.2328 \\
+ \textbf{Ours}
& \textbf{0.9855}
& \textbf{0.9547}
& \textbf{0.8957}
& \textbf{0.7178}
& \textbf{0.3777}
& \textbf{0.7002} \\
\hdashline
\rowcolor{oursbg}
\textit{$\Delta$ over VisualPrompter}
& +0.0344 & +0.1917 & +0.3789 & +0.1629 & +0.1456 & +0.4674 \\

\bottomrule
\end{tabular}
\end{table}

\begin{table}[tbp]
\centering
\caption{
\textbf{Efficiency comparison on GenEval2.}
We report generation quality, average inference time,
and the numbers of generator and LLM/VLM calls per prompt.
$\uparrow$ / $\downarrow$ indicate higher / lower is better.
}
\label{tab:efficiency}
\small
\setlength{\tabcolsep}{7pt}
\renewcommand{\arraystretch}{1.12}
\begin{adjustbox}{max width=\linewidth}
\begin{tabular}{lcccc}
\toprule
Method
& Score $\uparrow$
& Time (s) $\downarrow$
& Gen. Calls $\downarrow$
& LLM/VLM $\downarrow$ \\
\midrule
VisualPrompter & 0.2552 & 61.8  & 2.0  & 4.0 \\
Idea2Img       & 0.0586 & 453.2 & 12.0 & 8.0 \\
GenArtist      & 0.0556 & 117.3 & 3.2  & 4.6 \\
OctoT2I        & 0.2895 & 10.0  & 1.3  & 2.0 \\
\midrule
\rowcolor{oursbg}
\textbf{Ours}
& \textbf{0.7089}
& 42.6
& 1.4
& 2.1 \\
\bottomrule
\end{tabular}
\end{adjustbox}
\end{table}

\begin{figure}[tbp]
\centering
\includegraphics[width=0.75\linewidth]{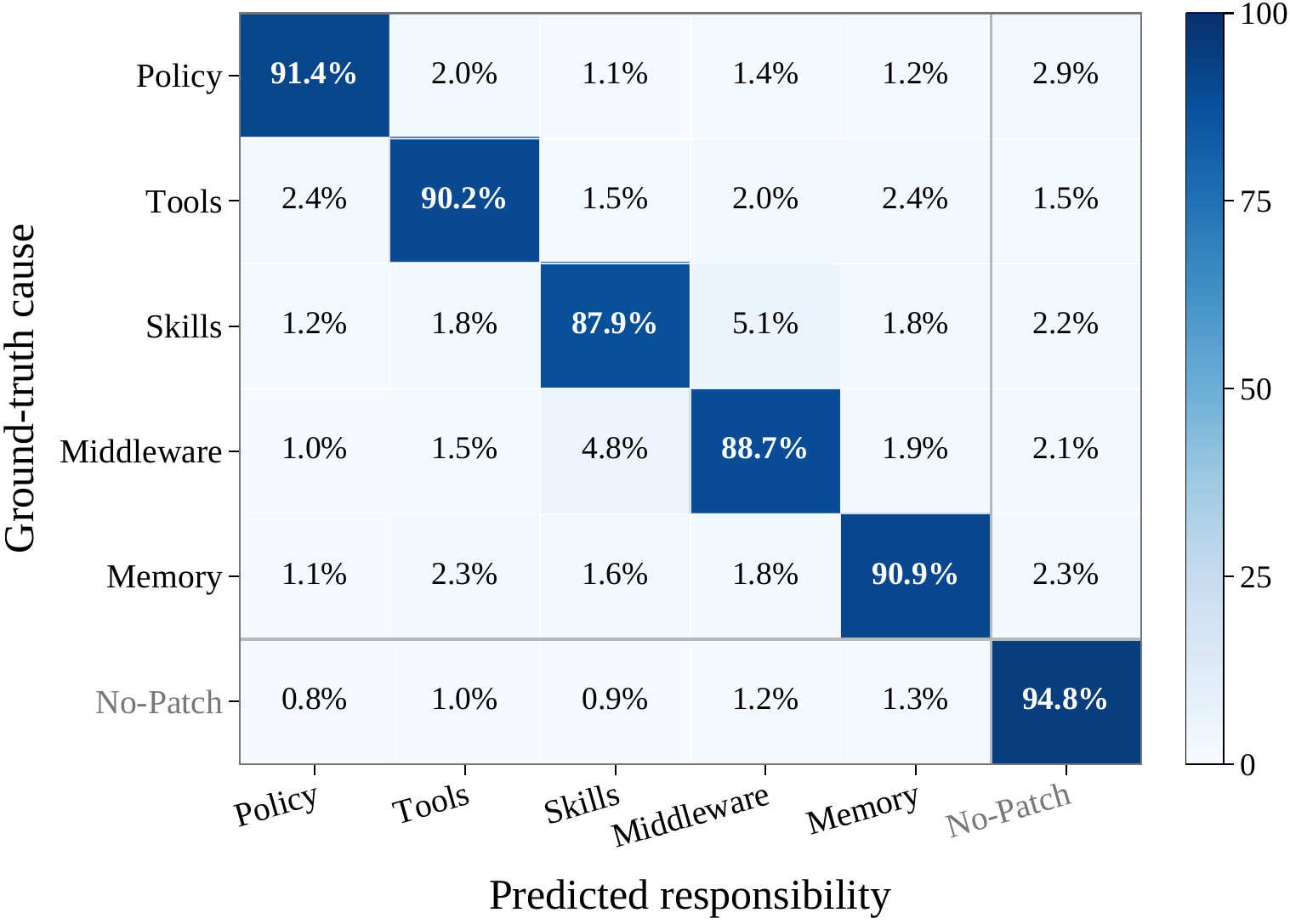}
\caption{\textbf{Harness attribution across persistent failure causes.}}
\label{fig:attribution}
\end{figure}

\subsection{Generalization and Efficiency}
\label{sec:backbone_generalization}

\paragraph{Generalization across frozen generators.}
A central goal of \textsc{EvoGen-Harness} is to improve the external system
without relying on the parameters of a particular image generator.
Table~\ref{tab:backbone_generalization} therefore applies the same framework to
three frozen backbones: FLUX.1-dev, Qwen-Image, and Janus-Pro.
Compared with VisualPrompter, \textsc{EvoGen-Harness} improves the overall
GenEval2 score by \textbf{+0.4537}, \textbf{+0.3750}, and
\textbf{+0.4674}, respectively.
The gain is positive for every reported visual skill on all three backbones.
This consistency supports the generator-agnostic nature of generator-external
harness evolution rather than an improvement tied to a particular T2I model. Additional qualitative cross-generator examples are provided in
Appendix~\ref{app:cross_generator_examples}.

\paragraph{Inference efficiency.}
Table~\ref{tab:efficiency} examines whether the performance gain simply comes
from additional sampling.
\textsc{EvoGen-Harness} uses 1.4 generator calls and 2.1 LLM/VLM calls per
prompt, close to OctoT2I's 1.3 and 2.0 calls, while achieving a substantially
higher GenEval2 score (\textbf{0.7089} vs.\ 0.2895).
OctoT2I remains faster in wall-clock latency
(10.0\,s vs.\ 42.6\,s), whereas \textsc{EvoGen-Harness} provides a markedly
higher generation score under a similar model-call budget.
Compared with VisualPrompter, our method is both faster
(42.6\,s vs.\ 61.8\,s) and substantially more accurate
(0.7089 vs.\ 0.2552).
Thus, the improvement cannot be explained by simply invoking more models or
sampling more images.

The search analysis in Appendix~\ref{app:search_efficiency} further supports
this result:
\textsc{Trace} retains a GenEval2 score of 0.7089, close to full-space search
(0.7112), while reducing the number of evaluated candidate edits from 20 to 8
and evolution-time generator calls from 80 to 32.

\subsection{Attribution and Evolution Analysis}
\label{sec:attribution_exp}

\paragraph{Harness localization.}
We evaluate the Harness Localizer on held-out controlled interventions, where
one responsibility is perturbed while the remaining factors are kept fixed.
As shown in Fig.~\ref{fig:attribution}, attribution recall reaches
\textbf{87.9--91.4\%} across the five editable responsibilities, while
\textsc{No-Patch} achieves \textbf{94.8\%}.
The strong diagonal structure indicates that trajectory-relative evidence
provides sufficient signal to distinguish different persistent failure sources.
Importantly, the high \textsc{No-Patch} accuracy shows that \textsc{Trace}
can also separate persistent harness weaknesses from stochastic generation
failures, avoiding unnecessary evolution.

\begin{figure}[tbp]
\centering
\includegraphics[width=0.87\linewidth]{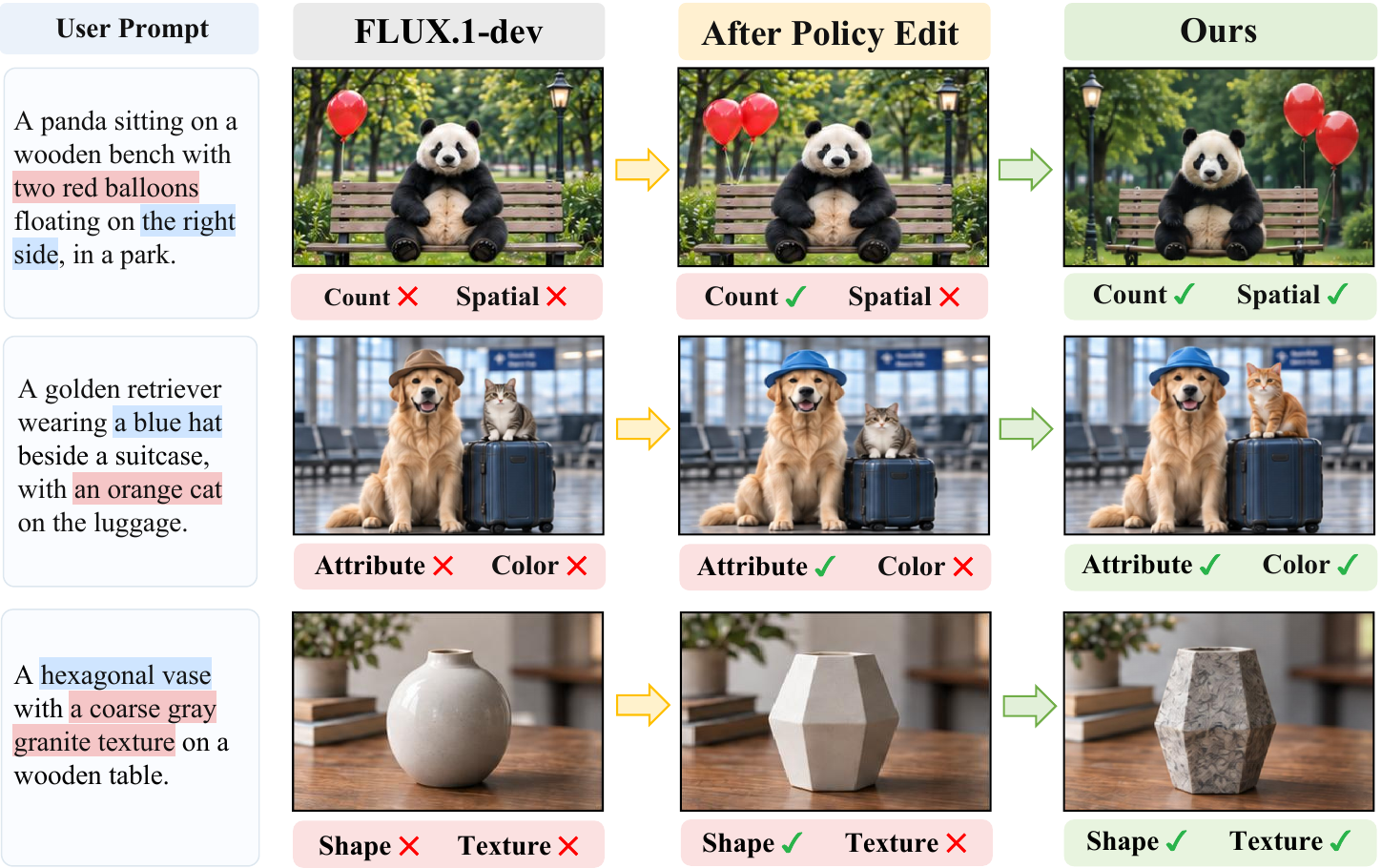}
\caption{
\textbf{Progressive visual repair with \textsc{Trace}.}
\textsc{Trace} progressively repairs multiple visual constraints through successive harness updates while keeping the generator frozen.
}
\label{fig:trace_evolution}
\end{figure}

\paragraph{Multi-responsibility evolution.}
Accurate localization is useful only if it leads to different adaptation
behaviors.
Table~\ref{tab:responsibility_analysis} shows that accepted updates span all
five responsibilities, with update shares of \textbf{13.0--27.8\%} and
positive held-out gains of \textbf{+0.089} to \textbf{+0.182}.
Policy is updated most frequently, whereas Skills yields the largest held-out
gain, suggesting that update frequency and transferable benefit capture
different aspects of harness evolution.
The update distribution therefore does not collapse to a single predefined
target; instead, the responsibility to evolve changes with the observed
failure.
Figure~\ref{fig:trace_evolution} further visualizes this progressive behavior:
successive harness updates resolve different constraints of the same failure,
while the underlying image generator remains frozen. Additional compound-failure results and responsibility-specific examples are
provided in Appendices~\ref{app:compound} and
\ref{app:responsibility_examples}.

\begin{table}[tbp]
\centering
\caption{
\textbf{Evolution across harness responsibilities.}
}
\label{tab:responsibility_analysis}
\small
\setlength{\tabcolsep}{7pt}
\renewcommand{\arraystretch}{1.12}
\begin{adjustbox}{max width=\linewidth}
\begin{tabular}{lcccc}
\toprule
Responsibility
& Attr. Recall $\uparrow$
& Update Share
& Commit Rate
& Held-out Gain $\uparrow$ \\
\midrule
Policy
& 91.4\% & 27.8\% & 74.2\% & +0.151 \\
Tools
& 90.2\% & 14.6\% & 69.8\% & +0.097 \\
Skills
& 87.9\% & 23.4\% & 76.5\% & +0.182 \\
Middleware
& 88.7\% & 21.2\% & 72.1\% & +0.139 \\
Memory
& 90.9\% & 13.0\% & 67.4\% & +0.089 \\
\bottomrule
\end{tabular}
\end{adjustbox}
\end{table}

\begin{table}[tbp]
\centering
\caption{
\textbf{Component ablation of \textsc{Trace}.}
}
\label{tab:trace_ablation}
\small
\setlength{\tabcolsep}{7pt}
\renewcommand{\arraystretch}{1.12}
\begin{adjustbox}{max width=\linewidth}
\begin{tabular}{lcccc}
\toprule
Variant
& GenEval2 $\uparrow$
& Attr. F1 $\uparrow$
& Held-out Gain $\uparrow$
& Regression $\downarrow$ \\
\midrule
w/o Traj. Aggregation
& 0.6580 & 0.816 & +0.111 & 5.1\% \\

w/o Attribution Guidance
& 0.6070 & 0.906 & +0.086 & 6.8\% \\

w/o Matched Evaluation
& 0.6390 & 0.906 & +0.101 & 7.6\% \\

w/o Progressive Re-local.
& 0.6250 & 0.906 & +0.091 & 4.4\% \\

w/o \textsc{No-Patch}
& 0.6420 & 0.753 & +0.103 & 10.7\% \\

w/o Safe Validation
& 0.6780 & 0.906 & +0.119 & 13.9\% \\

\midrule
\rowcolor{oursbg}
\textbf{\textsc{Trace} (Full)}
& \textbf{0.7089}
& \textbf{0.906}
& \textbf{+0.140}
& \textbf{1.9\%} \\
\bottomrule
\end{tabular}
\end{adjustbox}
\end{table}

\Needspace{8\baselineskip}
\subsection{Ablation Study}
\label{sec:ablation}

\paragraph{Evidence and guided evolution.}
As shown in Table~\ref{tab:trace_ablation}, removing trajectory aggregation
reduces GenEval2 from \textbf{0.7089} to 0.6580 and Attribution F1 from
\textbf{0.906} to 0.816, while unmatched evaluation increases regression to
7.6\%.
Removing attribution guidance causes the largest performance drop to 0.6070,
and disabling progressive re-localization further reduces the score to 0.6250.
These results show that reliable trajectory evidence is critical not only for
localizing persistent failures, but also for narrowing and redirecting the
subsequent evolution search.

\paragraph{Selective persistence.}
\textsc{No-Patch} and safe validation control which changes become persistent.
Removing them increases regression from \textbf{1.9\%} to 10.7\% and 13.9\%,
respectively.
The full \textsc{Trace} therefore achieves the strongest balance between
generation quality and persistent improvement, reaching
\textbf{0.7089} GenEval2 and \textbf{+0.140} held-out gain with only
\textbf{1.9\%} regression.
These results show that effective harness evolution depends on reliable
localization, progressive search, and selective retention working together. Additional sensitivity and proposal-LLM robustness results are provided in
Appendices~\ref{app:sensitivity} and \ref{app:llm_robustness}.




\FloatBarrier
\Needspace{13\baselineskip}
\section{Conclusion}

We introduce \textsc{EvoGen-Harness}, a generator-agnostic framework for
multi-responsibility evolution around frozen image generators.
Its core mechanism, \textsc{TRACE}, uses trajectory-relative visual evidence
to guide search over persistent harness responsibilities and progressively
coordinates updates through re-attribution and regression-safe validation.
Across GenEval2, T2I-CompBench++, and WISE, \textsc{EvoGen-Harness}
consistently improves generation quality while achieving reliable attribution,
low regression, and consistent improvements across frozen T2I backbones.
These results suggest that selectively evolving the external harness provides
a practical path toward more capable and self-improving image-generation
systems without modifying generator parameters.
\clearpage
\phantomsection
\addcontentsline{toc}{section}{References}
\bibliographystyle{plainnat}
\bibliography{references}

@inproceedings{wu2026reprompt,
  title={Reprompt: Reasoning-augmented reprompting for text-to-image generation via reinforcement learning},
  author={Wu, Mingrui and Wang, Lu and Zhao, Pu and Yang, Fangkai and Zhang, Jianjin and Liu, Jianfeng and Zhan, Yuefeng and Han, Weihao and Sun, Hao and Ji, Jiayi and others},
  booktitle={International Conference on Learning Representations},
  volume={2026},
  pages={14030--14057},
  year={2026}
}

@inproceedings{wu2026visualprompter,
  title={VisualPrompter: Semantic-Aware Prompt Optimization with Visual Feedback for Text-to-Image Synthesis},
  author={Wu, Shiyu and Sun, Mingzhen and Wang, Weining and Wang, Yequan and Liu, Jing},
  booktitle={International Conference on Learning Representations},
  volume={2026},
  pages={30578--30600},
  year={2026}
}

@inproceedings{jiang2026octot2i,
  author    = {Jiang, Xu and Chen, Bin and Li, Gehui and Duan, Yule and
               Wang, Ronggang and Zhang, Jian},
  title     = {OctoT2I: A Self-Evolving Agentic Text-to-Image Router},
  booktitle = {Proceedings of the IEEE/CVF Conference on Computer Vision and Pattern Recognition (CVPR)},
  month     = {June},
  year      = {2026},
  pages     = {31628--31638}
}

@misc{chen2026genrouter,
  title         = {GenRouter: Unified Workflow Routing for Agentic Image Generation},
  author        = {Chen, Harold Haodong and Hou, Zhiyu and Shu, Wen-Jie and
                   Ruan, Weilin and Xu, Yingjie and Guo, Litao and Chen, Ying-Cong},
  year          = {2026},
  eprint        = {2608.16721},
  archivePrefix = {arXiv},
  primaryClass  = {cs.CV}
}

@misc{li2026comfyclaw,
  title         = {{COMFYCLAW}: Self-Evolving Skill Harnesses for Image Generation Workflows},
  author        = {Li, Zongxia and Liu, Dawei and Liu, Fuxiao and Zhou, Yuhang and
                   Wu, Xiyang and Chen, Jingxi and Xie, Jing and Wu, Xiaomin and Sun, Lichao},
  year          = {2026},
  eprint        = {2607.01709},
  archivePrefix = {arXiv},
  primaryClass  = {cs.AI}
}

@misc{lin2026ahe,
  title         = {Agentic Harness Engineering: Observability-Driven Automatic Evolution of Coding-Agent Harnesses},
  author        = {Lin, Jiahang and Liu, Shichun and Pan, Chengjun and Lin, Lizhi and
                   Dou, Shihan and Xi, Zhiheng and Huang, Xuanjing and Yan, Hang and
                   Han, Zhenhua and Gui, Tao and Jiang, Yu-Gang},
  year          = {2026},
  eprint        = {2604.25850},
  archivePrefix = {arXiv},
  primaryClass  = {cs.CL}
}

@misc{zhang2026selfharness,
  title         = {Self-Harness: Harnesses That Improve Themselves},
  author        = {Zhang, Hangfan and Zhang, Shao and Li, Kangcong and Zhang, Chen and
                   Chen, Yang and Zhang, Yiqun and Bai, Lei and Hu, Shuyue},
  year          = {2026},
  eprint        = {2606.09498},
  archivePrefix = {arXiv},
  primaryClass  = {cs.AI}
}

@inproceedings{wang2024genartist,
  title     = {GenArtist: Multimodal LLM as an Agent for Unified Image Generation and Editing},
  author    = {Wang, Zhenyu and Li, Aoxue and Li, Zhenguo and Liu, Xihui},
  booktitle = {Advances in Neural Information Processing Systems},
  volume    = {37},
  pages     = {128374--128395},
  year      = {2024}
}

@inproceedings{yang2024idea2img,
  title     = {Idea2Img: Iterative Self-Refinement with GPT-4V for Automatic Image Design and Generation},
  author    = {Yang, Zhengyuan and Wang, Jianfeng and Li, Linjie and Lin, Kevin and
               Lin, Chung-Ching and Liu, Zicheng and Wang, Lijuan},
  booktitle = {European Conference on Computer Vision (ECCV)},
  pages     = {167--184},
  year      = {2024}
}

@inproceedings{ghosh2023geneval,
  title     = {GenEval: An Object-Focused Framework for Evaluating Text-to-Image Alignment},
  author    = {Ghosh, Dhruba and Hajishirzi, Hannaneh and Schmidt, Ludwig},
  booktitle = {Advances in Neural Information Processing Systems},
  volume    = {36},
  pages     = {52132--52152},
  year      = {2023}
}

@article{huang2025t2icompbenchpp,
  title={T2i-compbench++: An enhanced and comprehensive benchmark for compositional text-to-image generation},
  author={Huang, Kaiyi and Duan, Chengqi and Sun, Kaiyue and Xie, Enze and Li, Zhenguo and Liu, Xihui},
  journal={IEEE Transactions on Pattern Analysis and Machine Intelligence},
  volume={47},
  number={5},
  pages={3563--3579},
  year={2025},
  publisher={IEEE}
}

@inproceedings{cho2024dsg,
  title     = {Davidsonian Scene Graph: Improving Reliability in Fine-grained Evaluation for Text-to-Image Generation},
  author    = {Cho, Jaemin and Hu, Yushi and Baldridge, Jason and Garg, Roopal and
               Anderson, Peter and Krishna, Ranjay and Bansal, Mohit and
               Pont-Tuset, Jordi and Wang, Su},
  booktitle = {The Twelfth International Conference on Learning Representations},
  year      = {2024}
}

@inproceedings{rombach2022ldm,
  title={High-resolution image synthesis with latent diffusion models},
  author={Rombach, Robin and Blattmann, Andreas and Lorenz, Dominik and Esser, Patrick and Ommer, Bj{\"o}rn},
  booktitle={2022 IEEE/CVF conference on computer vision and pattern recognition (CVPR)},
  pages={10674--10685},
  year={2022},
  organization={ieee}
}

@inproceedings{esser2024sd3,
  title={Scaling rectified flow transformers for high-resolution image synthesis},
  author={Esser, Patrick and Kulal, Sumith and Blattmann, Andreas and Entezari, Rahim and M{\"u}ller, Jonas and Saini, Harry and Levi, Yam and Lorenz, Dominik and Sauer, Axel and Boesel, Frederic and others},
  booktitle={Forty-first international conference on machine learning},
  year={2024}
}

@inproceedings{zhao2024diffagent,
  title={Diffagent: Fast and accurate text-to-image api selection with large language model},
  author={Zhao, Lirui and Yang, Yue and Zhang, Kaipeng and Shao, Wenqi and Zhang, Yuxin and Qiao, Yu and Luo, Ping and Ji, Rongrong},
  booktitle={2024 IEEE/CVF Conference on Computer Vision and Pattern Recognition (CVPR)},
  pages={6390--6399},
  year={2024},
  organization={IEEE}
}

@inproceedings{hao2023promptist,
  title={Optimizing prompts for text-to-image generation},
  author={Hao, Yaru and Chi, Zewen and Dong, Li and Wei, Furu},
  journal={Advances in Neural Information Processing Systems},
  volume={36},
  pages={66923--66939},
  year={2023}
}

@misc{chen2026genevolve,
  title={GenEvolve: Self-Evolving Image Generation Agents via Tool-Orchestrated Visual Experience Distillation},
  author={Chen, Sixiang and Xing, Zhaohu and Ye, Tian and Geng, Xinyu and Lin, Yunlong and Lai, Jianyu and He, Xuanhua and Zhai, Fuxiang and Gao, Jialin and Zhu, Lei},
  journal={arXiv preprint arXiv:2605.21605},
  year={2026}
}

@inproceedings{cao2023beautifulprompt,
  title={Beautifulprompt: Towards automatic prompt engineering for text-to-image synthesis},
  author={Cao, Tingfeng and Wang, Chengyu and Liu, Bingyan and Wu, Ziheng and Zhu, Jinhui and Huang, Jun},
  booktitle={Proceedings of the 2023 Conference on Empirical Methods in Natural Language Processing: Industry Track},
  pages={1--11},
  year={2023}
}

@inproceedings{wu2024selfcorrecting,
  title={Self-correcting llm-controlled diffusion models},
  author={Wu, Tsung-Han and Lian, Long and Gonzalez, Joseph E and Li, Boyi and Darrell, Trevor},
  booktitle={2024 IEEE/CVF Conference on Computer Vision and Pattern Recognition (CVPR)},
  pages={6327--6336},
  year={2024},
  organization={IEEE}
}

@article{kamath2025geneval,
  title={Geneval 2: Addressing benchmark drift in text-to-image evaluation},
  author={Kamath, Amita and Chang, Kai-Wei and Krishna, Ranjay and Zettlemoyer, Luke and Hu, Yushi and Ghazvininejad, Marjan},
  journal={arXiv preprint arXiv:2512.16853},
  year={2025}
}

@inproceedings{niu2026wise,
  title={{WISE}: World Knowledge-Informed Semantic Evaluation for Text-to-Image Generation},
  author={Yuwei Niu and Munan Ning and Mengren Zheng and Weiyang Jin and Bin Lin and Peng Jin and Jiaqi Liao and Chaoran Feng and Fanqing Meng and Kun-Peng Ning and Bin Zhu and Li Yuan},
  booktitle={Forty-third International Conference on Machine Learning},
  year={2026},
}

@inproceedings{rosenman2024neuroprompts,
  title={Neuroprompts: An adaptive framework to optimize prompts for text-to-image generation},
  author={Rosenman, Shachar and Lal, Vasudev and Howard, Phillip},
  booktitle={Proceedings of the 18th Conference of the European Chapter of the Association for Computational Linguistics: System Demonstrations},
  pages={159--167},
  year={2024}
}

@inproceedings{jia2025chatgen,
  title={Chatgen: Automatic text-to-image generation from freestyle chatting},
  author={Jia, Chengyou and Xia, Changliang and Dang, Zhuohang and Wu, Weijia and Qian, Hangwei and Luo, Minnan},
  booktitle={2025 IEEE/CVF Conference on Computer Vision and Pattern Recognition (CVPR)},
  pages={13284--13293},
  year={2025},
  organization={IEEE}
}

@misc{flux2024,
  author = {{Black Forest Labs}},
  title  = {{FLUX}},
  year   = {2024},
  url    = {https://github.com/black-forest-labs/flux},
  note   = {GitHub repository}
}

@article{chen2025janus,
  title={Janus-pro: Unified multimodal understanding and generation with data and model scaling},
  author={Chen, Xiaokang and Wu, Zhiyu and Liu, Xingchao and Pan, Zizheng and Liu, Wen and Xie, Zhenda and Yu, Xingkai and Ruan, Chong},
  journal={arXiv preprint arXiv:2501.17811},
  year={2025}
}

@inproceedings{liu2025flow,
  title={Flow-grpo: Training flow matching models via online rl},
  author={Liu, Jie and Liu, Gongye and Liang, Jiajun and Li, Yangguang and Liu, Jiaheng and Wang, Xintao and Wan, Pengfei and Zhang, Di and Ouyang, Wanli},
  journal={Advances in neural information processing systems},
  volume={38},
  pages={40783--40818},
  year={2026}
}

@misc{wu2025qwenimage,
  title={Qwen-image technical report},
  author={Wu, Chenfei and Li, Jiahao and Zhou, Jingren and Lin, Junyang and Gao, Kaiyuan and Yan, Kun and Yin, Sheng-ming and Bai, Shuai and Xu, Xiao and Chen, Yilei and others},
  journal={arXiv preprint arXiv:2508.02324},
  year={2025}
}

@article{team2025zimage,
  title={Z-image: An efficient image generation foundation model with single-stream diffusion transformer},
  author={Cai, Huanqia and Cao, Sihan and Du, Ruoyi and Gao, Peng and Hao, Aiming and Hoi, Steven and Hou, Zhaohui and Huang, Shijie and Jiang, Dengyang and Jiang, Yuming and others},
  journal={arXiv preprint arXiv:2511.22699},
  year={2025}
}

@misc{google2025nanobanana,
  author       = {{Google}},
  title        = {{Gemini 2.5 Flash Image (Nano Banana)}},
  year         = {2025},
  howpublished = {Google AI for Developers},
  url          = {https://ai.google.dev/gemini-api/docs/models/gemini-2.5-flash-image}
}

@inproceedings{feng2023layoutgpt,
  title={Layoutgpt: Compositional visual planning and generation with large language models},
  author={Feng, Weixi and Zhu, Wanrong and Fu, Tsu-jui and Jampani, Varun and Akula, Arjun and He, Xuehai and Basu, Sugato and Wang, Xin Eric and Wang, William Yang},
  journal={Advances in neural information processing systems},
  volume={36},
  pages={18225--18250},
  year={2023}
}

@article{lian2023llmgrounded,
  title={Llm-grounded diffusion: Enhancing prompt understanding of text-to-image diffusion models with large language models},
  author={Lian, Long and Li, Boyi and Yala, Adam and Darrell, Trevor},
  journal={arXiv preprint arXiv:2305.13655},
  year={2023}
}

@inproceedings{yang2024mastering,
  title={Mastering text-to-image diffusion: Recaptioning, planning, and generating with multimodal llms},
  author={Yang, Ling and Yu, Zhaochen and Meng, Chenlin and Xu, Minkai and Ermon, Stefano and Cui, Bin},
  booktitle={Forty-first International Conference on Machine Learning},
  year={2024}
}

@article{wang2024divide,
  title={Divide and conquer: Language models can plan and self-correct for compositional text-to-image generation},
  author={Wang, Zhenyu and Xie, Enze and Li, Aoxue and Wang, Zhongdao and Liu, Xihui and Li, Zhenguo},
  journal={arXiv preprint arXiv:2401.15688},
  year={2024}
}

@inproceedings{feng2024ranni,
  title={Ranni: Taming text-to-image diffusion for accurate instruction following},
  author={Feng, Yutong and Gong, Biao and Chen, Di and Shen, Yujun and Liu, Yu and Zhou, Jingren},
  booktitle={2024 IEEE/CVF Conference on Computer Vision and Pattern Recognition (CVPR)},
  pages={4744--4753},
  year={2024},
  organization={IEEE}
}

@inproceedings{zhang2025itercomp,
  title={Itercomp: Iterative composition-aware feedback learning from model gallery for text-to-image generation},
  author={Zhang, Xinchen and Yang, Ling and Li, Guohao and Cai, Yaqi and Tang, Yong and Yang, Yujiu and Wang, Mengdi and CUI, Bin and others},
  booktitle={International Conference on Learning Representations},
  volume={2025},
  pages={31968--31988},
  year={2025}
}

@article{sun2025marmot,
  title={Marmot: Multi-agent reasoning for multi-object self-correcting in improving image-text alignment},
  author={Sun, Jiayang and Wang, Hongbo and Cao, Jie and Huang, Huaibo and He, Ran},
  journal={arXiv e-prints},
  pages={arXiv--2504},
  year={2025}
}

@inproceedings{li2025mccd,
  title={Mccd: Multi-agent collaboration-based compositional diffusion for complex text-to-image generation},
  author={Li, Mingcheng and Hou, Xiaolu and Liu, Ziyang and Yang, Dingkang and Qian, Ziyun and Chen, Jiawei and Wei, Jinjie and Jiang, Yue and Xu, Qingyao and Zhang, Lihua},
  booktitle={2025 IEEE/CVF Conference on Computer Vision and Pattern Recognition (CVPR)},
  pages={13263--13272},
  year={2025},
  organization={IEEE}
}

@article{yang2026m3,
  title={M3: High-fidelity Text-to-Image Generation via Multi-Modal, Multi-Agent and Multi-Round Visual Reasoning},
  author={Yang, Bangji and Guo, Ruihan and Fan, Jiajun and Cheng, Chaoran and Liu, Ge},
  journal={arXiv preprint arXiv:2602.06166},
  year={2026}
}

@article{zhang2026jitagent,
  title={JIT-Agent: Scaling Harness Intelligence via Just-in-Time Harness Evolution},
  author={Zhang, Guibin and Lu, Leo and Xie, Fangzhou and Zhu, Kang and Wang, Junhao and Xie, Zhifei and Yu, Zhaochen and Liu, Zihang and Sun, Zhongxiang and Li, Qiankun and others},
  journal={arXiv preprint arXiv:2608.25593},
  year={2026}
}
\clearpage
\appendix
\section{Evaluation Metrics}
\label{sec:appendix}
\label{app:metrics}

We follow the official metrics and evaluation protocols for
GenEval2, T2I-CompBench++, and WISE.
Here we define only the auxiliary metrics used to analyze
\textsc{Trace}'s attribution and persistent evolution.

\paragraph{Attribution metrics.}
Let
$\mathcal{R}=
\{\Pi,\mathcal{T},\mathcal{S},\mathcal{W},\mathcal{M}\}$
denote the five editable harness responsibilities.
For each responsibility $z\in\mathcal{R}$, we compute precision and recall as
\begin{equation}
    P_z
    =
    \frac{\mathrm{TP}_z}
         {\mathrm{TP}_z+\mathrm{FP}_z},
    \qquad
    R_z
    =
    \frac{\mathrm{TP}_z}
         {\mathrm{TP}_z+\mathrm{FN}_z}.
\end{equation}
We report macro-averaged attribution F1 over the editable responsibilities:
\begin{equation}
    \mathrm{F1}_{\mathrm{attr}}
    =
    \frac{1}{|\mathcal{R}|}
    \sum_{z\in\mathcal{R}}
    \frac{2P_zR_z}{P_z+R_z}.
\end{equation}
For the class-wise attribution results in Fig.~\ref{fig:attribution},
we additionally report the fraction of examples from each ground-truth
responsibility that are correctly attributed.

The abstention class $\varnothing$ is evaluated separately.
Its \textsc{No-Patch} accuracy is
\begin{equation}
    \mathrm{Acc}_{\mathrm{NP}}
    =
    \frac{\mathrm{TP}_{\varnothing}}
         {N_{\varnothing}},
\end{equation}
where $N_{\varnothing}$ is the number of ground-truth
\textsc{No-Patch} examples.
This corresponds to the diagonal entry of the \textsc{No-Patch} row in the
row-normalized confusion matrix.

\paragraph{Held-out gain.}
To measure whether an accepted update transfers beyond the examples that
triggered it, we compare the evolved harness with its parent harness on a
disjoint held-out set $\mathcal{D}_{\mathrm{held}}$.
For matched stochastic conditions
$\Omega=\{\omega_1,\ldots,\omega_K\}$, define
\begin{equation}
    \bar{Q}(x;\mathcal{H})
    =
    \frac{1}{K}
    \sum_{k=1}^{K}
    Q(x;\mathcal{H},\omega_k),
\end{equation}
where $Q$ is the corresponding visual evaluation score.
Held-out Gain is then
\begin{equation}
    \mathrm{Gain}_{\mathrm{held}}
    =
    \frac{1}{|\mathcal{D}_{\mathrm{held}}|}
    \sum_{x\in\mathcal{D}_{\mathrm{held}}}
    \left[
        \bar{Q}(x;\mathcal{H}_{t+1})
        -
        \bar{Q}(x;\mathcal{H}_{t})
    \right].
\end{equation}
A positive value therefore indicates that the accepted update improves
previously unseen cases rather than only repairing the triggering example.

\paragraph{Regression.}
We measure regression on a preservation set
$\mathcal{D}_{\mathrm{pres}}$ containing cases that are successfully handled
before evolution.
Let
\begin{equation}
    S(x;\mathcal{H})
    =
    \mathbb{I}
    \left[
        \bar{Q}(x;\mathcal{H}) \geq \theta_x
    \right]
\end{equation}
denote whether harness $\mathcal{H}$ successfully handles case $x$ under the
corresponding evaluation threshold $\theta_x$.
Defining
\[
    \mathcal{D}_{\mathrm{pres}}^{+}
    =
    \left\{
        x\in\mathcal{D}_{\mathrm{pres}}
        :
        S(x;\mathcal{H}_{t})=1
    \right\},
\]
the regression rate is
\begin{equation}
    \mathrm{Regression}
    =
    \frac{100}
         {|\mathcal{D}_{\mathrm{pres}}^{+}|}
    \sum_{x\in\mathcal{D}_{\mathrm{pres}}^{+}}
    \mathbb{I}
    \left[
        S(x;\mathcal{H}_{t+1})=0
    \right].
\end{equation}
Thus, regression measures the percentage of previously successful cases that
become unsuccessful after a persistent harness update.

\paragraph{Responsibility statistics.}
For Table~\ref{tab:responsibility_analysis},
\emph{Update Share} is the fraction of all committed updates assigned to a
given responsibility, while \emph{Commit Rate} is the fraction of proposed
updates for that responsibility that pass validation and are retained.
These statistics describe how evolution is distributed across the harness and
are not treated as optimization objectives.

\paragraph{Efficiency metrics.}
We report average wall-clock inference time per prompt together with the
numbers of generator and LLM/VLM calls.
For the search-efficiency analysis, Candidate Edits denotes the average number
of candidate harness modifications actually evaluated per task.

\section{Implementation Details}
\label{app:implementation}

\paragraph{Benchmark and evaluation settings.}
We use the official releases and evaluation pipelines for all three
benchmarks. For GenEval2, we evaluate on the official 800 prompts and
report Soft-TIFA following the official aggregation protocol.
For T2I-CompBench++, we follow the official evaluation setup, including
the prescribed sampling and fixed-seed protocol, and report the eight
official compositional metrics.
For WISE, we use the latest verified benchmark release and its official
WiScore evaluation pipeline.

Benchmark evaluation is separated from the stochastic executions used
internally by \textsc{Trace}. Unless otherwise specified,
\textsc{Trace} uses $K=4$ stochastic executions for trajectory-relative
evidence collection. Benchmark labels, official scores, and test
annotations are never exposed to the harness localizer, edit proposer,
or candidate-selection procedure.

\paragraph{System configuration.}
Unless otherwise specified, we use FLUX.1-dev as the frozen image generator
and GPT-4.1 as the fixed LLM for both harness localization and
responsibility-conditioned edit proposal. For fair comparison, all baseline
methods are evaluated using their official implementations and recommended
default settings.
The Harness Localizer and Edit Proposer use separate prompts and structured
output schemas. OWLv2-base with NVILA-Lite-2B-Verifier provides visual
verification, while a deterministic execution controller executes the evolved harness
during deployment, coordinating requirement parsing, tool invocation,
generation procedures, and verification.
All models and controllers remain fixed throughout evolution; only the
persistent generator-external harness is updated.
Benchmark labels, official benchmark scores, and test annotations are never
provided to the localization, proposal, or candidate-selection procedure.

\paragraph{\textsc{Trace} configuration.}
Unless otherwise specified, we use an attribution frontier size of $r=3$,
$K=4$ stochastic executions for trajectory-relative evidence,
at most $M=3$ candidate edits per responsibility, and a maximum evolution
depth of $L=3$. We set the preservation and execution-cost coefficients to
$\lambda = 0.5$ and $\mu = 0.05$, respectively.
The Harness Localizer ranks the five editable responsibilities together with
\textsc{No-Patch} directly from trajectory-relative evidence and requires no
additional training. 
All candidate edits are evaluated against their parent harness under matched
stochastic conditions. 

\paragraph{Validation and compute.}
Target and held-out validation use at most three examples each, while the
preservation set contains at most two examples.
We use a regression tolerance of $\epsilon=0.01$ and at least two validation
seeds.
The evolution loop allows up to two proposal attempts when an edit is invalid
or cannot be executed.
All main experiments are conducted on 2 NVIDIA A100 (80\,GB) GPUs.
Reported LLM/VLM calls include harness-localization, edit-proposal, and
visual-verification calls.

\paragraph{Evolution data and evaluation separation.}
\textsc{EvoGen-Harness} performs no additional parameter training; all image
generators, the harness localizer, the edit proposer, visual verifiers, and the
deterministic controller remain fixed throughout evolution, and only the
persistent generator-external harness is updated. We construct a dedicated
evolution corpus from PartiPrompts (P2), independent of the benchmarks used for
final evaluation. After de-duplication and benchmark-overlap filtering, we form
three pairwise-disjoint subsets: an evolution set
$\mathcal{D}_{\mathrm{evo}}$ containing $500$ prompts, a held-out validation
set $\mathcal{D}_{\mathrm{val}}$ containing $100$ prompts, and a preservation
set $\mathcal{D}_{\mathrm{pres}}$ containing $100$ prompts. During each
evolution step, at most three examples from $\mathcal{D}_{\mathrm{val}}$ and
at most two examples from $\mathcal{D}_{\mathrm{pres}}$ are used for held-out
and preservation validation, respectively.

To prevent evaluation contamination, we remove exact and near-duplicate prompts
overlapping with the official evaluation prompts of GenEval2,
T2I-CompBench++, and WISE before constructing these splits. The evolution,
held-out validation, and preservation sets are therefore disjoint from all final
benchmark evaluation sets. Benchmark test prompts, labels, annotations, and
official evaluation scores are never exposed to harness localization, edit
proposal, candidate selection, or validation. After evolution, the resulting
harness is frozen and used unchanged for final benchmark evaluation; no
benchmark test prompt is used to update the harness or can influence subsequent
test examples.

\paragraph{Harness representations and edit operators.}
The five harness responsibilities are instantiated as explicit and
independently editable persistent records. Their concrete roles and editable
contents are summarized in Table~\ref{tab:harness_schema}. This decomposition
makes the heterogeneous harness state operational: an edit is localized to one
responsibility and cannot directly modify the remaining responsibilities or the
frozen generation system.

\begin{table}[t]
\centering
\small
\caption{Operational representation of the five persistent harness
responsibilities. The examples describe the editable state rather than the
parameters of the frozen generator.}
\label{tab:harness_schema}
\begin{tabular}{P{0.15\linewidth} P{0.31\linewidth} P{0.45\linewidth}}
\toprule
\textbf{Responsibility} & \textbf{Persistent representation}
& \textbf{Editable contents} \\
\midrule
Policy
& Visual requirement and constraint rules
& Requirement interpretation, constraint priority, and criteria used to
represent explicit visual requirements such as count, attribute, relation,
and action constraints. \\

Tools
& Capability registry for available generation and auxiliary tools
& Capability and limitation descriptions, applicable scenarios, and
tool-specific knowledge used by the harness. Tool implementations and model
parameters are never modified. \\

Skills
& Reusable procedural templates
& Multi-step generation routines, prompt templates, decomposition strategies,
and reusable task-solving procedures. \\

Middleware
& Run-time orchestration configuration
& Routing, verification, retry, termination, and tool-invocation rules.
The deterministic base controller itself remains fixed. \\

Memory
& Persistent cross-task experience records
& Reusable summaries of successful and failed executions, recurring failure
patterns, and previously validated corrective experience. \\
\bottomrule
\end{tabular}
\end{table}

Formally, each candidate edit can be represented at the harness interface as
\begin{equation}
\delta = (z, o, k, v_{\mathrm{old}}, v_{\mathrm{new}})
\end{equation}
where $z \in \{\Pi,T,S,W,M\}$ denotes the selected responsibility,
$o \in \{\textsc{Add},\textsc{Replace},\textsc{Delete}\}$ is the edit
operation, $k$ identifies the affected record or field, and
$v_{\mathrm{old}}$ and $v_{\mathrm{new}}$ denote its states before and after
editing. Applying an edit yields
\begin{equation}
H' = H \oplus \delta
\end{equation}
where only the state associated with $z$ is changed. Edits that modify another
responsibility, the frozen generator, or the deterministic base controller are
rejected as invalid.

The responsibility-conditioned proposer receives the current value of the
selected responsibility, the trajectory-relative evidence, and its permitted
edit schema. It proposes localized modifications only within this schema.
\textsc{NO-PATCH} is an attribution decision rather than a persistent harness
component and therefore does not modify $H$.

\paragraph{Examples of localized harness edits.}
The following examples illustrate how an observed visual failure is converted
into a responsibility-specific persistent modification.

\textbf{Example 1: Policy edit for an exact-count failure.}
Suppose the request requires ``three white ducks'', but repeated executions
produce only two ducks. After trajectory-relative attribution selects Policy,
the proposer may revise the corresponding requirement rule as follows:

\begin{quote}
\small
\textbf{Before:}
Explicit numerical descriptions are represented together with other object
attributes during requirement parsing.

\medskip
\textbf{After:}
Explicit cardinality expressions (e.g., ``three ducks'') must be represented as
hard count constraints. The required cardinality is preserved during prompt
construction and explicitly checked during visual verification.
\end{quote}

This edit changes only the persistent interpretation of numerical requirements;
the image generator, tool inventory, and execution controller remain unchanged.

\textbf{Example 2: Middleware edit for repeated verification failure.}
Consider a case in which the generated image violates a spatial or counting
constraint and the current execution terminates immediately after verification.
If the residual failure is attributed to Middleware, a possible edit is:

\begin{quote}
\small
\textbf{Before:}
Terminate the current generation procedure after the first verification pass.

\medskip
\textbf{After:}
When verification detects an unresolved explicit constraint and the execution
budget remains available, trigger one targeted retry using the verifier feedback.
Terminate when all required constraints pass verification or when the retry
budget is exhausted.
\end{quote}

The edit modifies the orchestration policy without changing the verifier,
generator, or underlying controller implementation.

Other responsibilities are edited through the same localized interface.
For example, a \textbf{Tools} edit may correct an inaccurate capability record
when repeated evidence shows that a generator is unreliable for exact-count
requests; a \textbf{Skills} edit may replace a direct-generation routine with a
validated procedure that first makes spatial relations explicit and then
verifies them after generation; and a \textbf{Memory} edit may store a
cross-task experience stating that explicit cardinality preservation and
post-generation counting consistently repaired a recurring family of failures.
All such edits are retained only after the matched and held-out validation
described in Sec.~\ref{sec:trace}.

\subsection{\textsc{Trace}: Progressive Harness Evolution}
\label{app:trace}

Algorithm~\ref{alg:trace} summarizes the complete \textsc{Trace} procedure.
Trajectory-relative localization first narrows the heterogeneous harness space
to a small set of plausible responsibilities.
A fixed LLM then proposes localized edits only within this frontier, and
matched execution feedback determines which edit should be applied.
After each accepted edit, residual failures are re-localized, allowing the
evolution path to move across responsibilities when necessary.

\begin{algorithm}[t]
\caption{\textsc{Trace}: Trajectory-Relative Attribution and Coordinated Evolution}
\label{alg:trace}
\small
\begin{algorithmic}[1]

\Require Current harness $\mathcal{H}_t$,
         frozen generator $\mathcal{G}$,
         frontier size $r$,
         stochastic executions $K$,
         candidate budget $M$,
         maximum depth $L$

\State $\widetilde{\mathcal{H}} \gets \mathcal{H}_t$
\State $\boldsymbol{\pi} \gets [\,]$

\For{$\ell=1,\ldots,L$}

    \State Collect trajectory-relative evidence
    $\mathcal{E}_{\ell}$ from $K$ stochastic executions
    under $\widetilde{\mathcal{H}}$

    \State Obtain localization scores
    $s_{\mathrm{loc}}
    (z\mid\mathcal{E}_{\ell},\widetilde{\mathcal{H}})$
    from the LLM-based Harness Localizer

    \If{\textsc{No-Patch} is preferred}
        \State \textbf{break}
    \EndIf

    \State
    $\mathcal{R}_{\ell}^{(r)}
    \gets
    \operatorname{Top}\text{-}r_{z\in\mathcal{R}}
    s_{\mathrm{loc}}
    (z\mid\mathcal{E}_{\ell},\widetilde{\mathcal{H}})$

    \State Generate at most $M$ localized edits
    $\Delta_z$ for each
    $z\in\mathcal{R}_{\ell}^{(r)}$
    using the LLM proposer

    \State
    $\Delta_{\ell}
    \gets
    \bigcup_{z\in\mathcal{R}_{\ell}^{(r)}}\Delta_z$

    \For{each $\delta\in\Delta_{\ell}$}
        \State Evaluate
        $\Delta Q_{\ell}(\delta)$
        under the same $K$ stochastic conditions
        \State Compute
        $U_{\ell}(\delta)$
    \EndFor

    \State
    $\delta_{\ell}^{*}
    \gets
    \arg\max_{\delta\in\Delta_{\ell}}
    U_{\ell}(\delta)$

    \If{$U_{\ell}(\delta_{\ell}^{*})\leq 0$}
        \State \textbf{break}
    \EndIf

    \State
    $\widetilde{\mathcal{H}}
    \gets
    \widetilde{\mathcal{H}}
    \oplus
    \delta_{\ell}^{*}$

    \State
    $\boldsymbol{\pi}
    \gets
    \boldsymbol{\pi}
    \mathbin{\|}\delta_{\ell}^{*}$

\EndFor

\If{$\widetilde{\mathcal{H}}$ passes held-out and preservation validation}
    \State \Return $\widetilde{\mathcal{H}}$
\Else
    \State \Return $\mathcal{H}_t$
\EndIf

\end{algorithmic}
\end{algorithm}

\paragraph{Search complexity.}
At each evolution step, the Harness Localizer retains at most $r$
responsibilities, each contributing at most $M$ candidate edits.
Therefore, \textsc{Trace} evaluates at most
\begin{equation}
    \mathcal{O}(LrM)
\end{equation}
candidate edits over $L$ evolution steps.
Since each candidate is evaluated under $K$ matched stochastic conditions,
the corresponding generation complexity is
\begin{equation}
    \mathcal{O}(LrMK).
\end{equation}
This replaces combinatorial enumeration over multi-responsibility edit
sequences with a bounded attribution-guided search, while progressive
re-localization still allows the evolution path to traverse different
responsibilities.

\subsection{Search Efficiency Analysis}
\label{app:search_efficiency}

We examine whether \textsc{Trace} reduces unnecessary exploration while
preserving generation quality.
We compare three search strategies under the same frozen generator,
LLM proposer, candidate budget, stochastic execution budget, maximum
evolution depth, and matched-evaluation protocol.
\emph{Full-Space Search} explores all editable responsibilities at each step;
\emph{Hard Top-1} explores only the responsibility with the highest
localization score; and \textsc{Trace} retains a Top-$r$ localization frontier
and progressively re-localizes residual failures after each accepted update.

\begin{table}[t]
\centering
\caption{
\textbf{Search efficiency of \textsc{Trace}.}
Top-$r$ localization substantially reduces exploration relative to full-space
search while preserving generation quality.
Candidate edits and generator calls are averaged per task.
$\uparrow$ / $\downarrow$ indicate higher / lower is better.
}
\label{tab:search_efficiency}

\vspace{0.3em}
\small
\setlength{\tabcolsep}{5.0pt}
\renewcommand{\arraystretch}{1.12}

\resizebox{\linewidth}{!}{
\begin{tabular}{lcccc}
\toprule
Search Strategy
& GenEval2 $\uparrow$
& Candidate Edits $\downarrow$
& Gen. Calls $\downarrow$
& Evol. Time (s) $\downarrow$ \\
\midrule

Full-Space Search
& 0.7112
& 20.0
& 80.0
& 2386 \\

Hard Top-1
& 0.6815
& \textbf{4.0}
& \textbf{16.0}
& \textbf{486} \\

\rowcolor{oursbg}
\textbf{\textsc{Trace} Top-$r$}
& 0.7089
& 8.0
& 32.0
& 964 \\

\bottomrule
\end{tabular}
}

\vspace{-0.4em}
\end{table}

Full-space search achieves only a marginally higher score than
\textsc{Trace} (0.7112 vs.\ 0.7089), but requires substantially more
candidate evaluations and generator calls.
Hard Top-1 is the most efficient strategy, yet its lower score (0.6815)
shows the risk of committing prematurely to a single responsibility.
In contrast, \textsc{Trace} preserves multiple plausible responsibilities in
a compact Top-$r$ frontier, providing a favorable balance between search
efficiency and generation quality while allowing subsequent evolution to be
redirected through progressive re-localization.

\paragraph{Evolution-time cost accounting.}
We distinguish deployment-time inference cost from the one-time cost of
harness evolution. Table~\ref{tab:efficiency} reports the former, i.e., the average cost of
executing the current harness for a test request. In contrast,
Table~\ref{tab:search_efficiency} reports the computation incurred by
\textsc{Trace} when a persistent update is triggered, including candidate
evaluation under matched stochastic conditions.

All search-efficiency statistics are averaged per evolution task.
The reported generator calls include candidate evaluation rollouts, while
the evolution time covers localization, proposal, candidate evaluation,
and validation. Since accepted harness updates persist across subsequent
requests, this evolution cost is amortized over future executions and is
not included in the per-prompt inference cost reported in Table~\ref{tab:efficiency}.

\subsection{Robustness to Compound Failures}
\label{app:compound}

The attribution analysis in Sec.~\ref{sec:attribution_exp} uses controlled
single-responsibility interventions to obtain unambiguous ground-truth
failure causes. In realistic generation pipelines, however, a failure
may arise from multiple harness responsibilities simultaneously.
We therefore conduct an additional controlled stress test to evaluate
the robustness of \textsc{Trace} under compound failure causes.

\paragraph{Paired-responsibility interventions.}
We construct compound failures by simultaneously injecting two
independently defined persistent perturbations into different harness
responsibilities. We evaluate all ten unordered pairs among
$\{\Pi,T,S,W,M\}$. For each compound case, the request, frozen
generator, stochastic conditions, and all remaining harness
responsibilities are kept unchanged. This construction isolates the
effect of interacting persistent causes while preserving exact
ground-truth attribution.

Let $Z_i^\star=\{z_i^{(1)},z_i^{(2)}\}$ denote the two ground-truth
responsibilities for example $i$, and let $\mathcal{R}_i^{(3)}$ denote
the initial Top-3 localization frontier. We measure initial localization
quality using
\begin{equation}
\mathrm{CauseRecall@3}
=
\frac{1}{N}
\sum_{i=1}^{N}
\frac{
    \left|
    \mathcal{R}_i^{(3)} \cap Z_i^\star
    \right|
}{
    \left|Z_i^\star\right|
}.
\end{equation}
Because \textsc{Trace} re-localizes residual failures after each
accepted edit, we additionally report \textbf{Both Recovered}, the
fraction of cases for which both ground-truth responsibilities are
identified over the complete progressive evolution path.
Finally, \textbf{Repair Success} measures the fraction of compound
cases for which both injected failure effects are successfully resolved
after evolution.

\begin{table}[t]
\centering
\caption{
Robustness of \textsc{Trace} under paired-responsibility failures.
Each example contains two simultaneously injected persistent causes.
Cause Recall@3 measures coverage of the two ground-truth causes by the
initial Top-3 localization frontier; Both Recovered measures whether
both causes are identified over progressive re-localization; and Repair
Success measures whether both injected failure effects are resolved
after evolution. $\uparrow$ indicates higher is better.
}
\label{tab:compound_causes}
\small
\setlength{\tabcolsep}{5.5pt}
\begin{tabular}{lccc}
\toprule
\textbf{Compound Causes}
& \textbf{Cause Recall@3 $\uparrow$}
& \textbf{Both Recovered $\uparrow$}
& \textbf{Repair Success $\uparrow$} \\
\midrule
Policy + Tools              & 96.1\% & 92.4\% & 86.8\% \\
Policy + Skills             & 94.8\% & 90.1\% & 85.7\% \\
Policy + Middleware         & 94.6\% & 89.5\% & 84.3\% \\
Policy + Memory             & 95.5\% & 91.3\% & 85.1\% \\
Tools + Skills              & 94.1\% & 88.7\% & 83.6\% \\
Tools + Middleware          & 93.8\% & 87.9\% & 82.4\% \\
Tools + Memory              & 95.0\% & 90.4\% & 83.2\% \\
Skills + Middleware         & 91.9\% & 83.8\% & 79.6\% \\
Skills + Memory             & 93.6\% & 87.1\% & 82.8\% \\
Middleware + Memory         & 94.0\% & 88.2\% & 81.9\% \\
\midrule
\rowcolor{oursbg}
\textbf{Average}
& \textbf{94.3\%}
& \textbf{88.9\%}
& \textbf{83.5\%} \\
\bottomrule
\end{tabular}
\end{table}

\paragraph{Results.}
As shown in Table~\ref{tab:compound_causes}, \textsc{Trace} remains
robust when two persistent failure causes are present simultaneously.
The initial Top-3 frontier covers 94.3\% of ground-truth causes on
average, indicating that attribution remains reliable despite competing
evidence from multiple responsibilities. More importantly, progressive
re-localization recovers both causes in 88.9\% of compound cases,
showing that \textsc{Trace} does not require an entire generation
failure to originate from a single responsibility.

The final repair success rate reaches 83.5\%. The gap between cause
recovery and successful repair indicates that correctly identifying
both responsibilities does not always guarantee that the proposed edits
fully eliminate the corresponding visual failures. The most challenging
combination is Skills + Middleware, with a repair success rate of
79.6\%, suggesting stronger interaction between reusable procedures and
run-time orchestration. Nevertheless, performance remains consistent
across all ten responsibility pairs, supporting the ability of
progressive re-localization to handle compound persistent failures.

Importantly, this compound-failure setting directly tests the motivation
for coordinated multi-responsibility evolution: when multiple persistent
failure sources coexist, effective evolution must remain open to different
intervention sites rather than committing to a single predefined
responsibility. The strong \emph{Both Recovered} and \emph{Repair Success}
results show that progressive re-localization enables \textsc{Trace} to
redirect evolution across responsibilities as residual failures change,
rather than collapsing to a fixed adaptation target.

\subsection{Sensitivity Analysis}
\label{app:sensitivity}

We study the sensitivity of \textsc{Trace} to three search parameters:
the attribution frontier size $r$, the number of stochastic executions $K$,
and the maximum evolution depth $L$.
All remaining settings follow Appendix~\ref{app:implementation}.

\begin{figure}[t]
    \centering
    \includegraphics[width=0.98\textwidth]
    {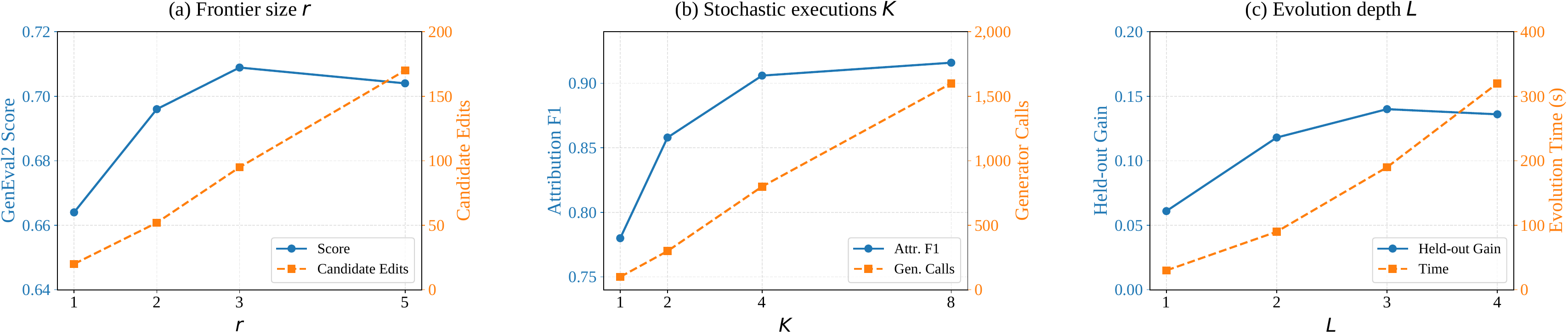}
    \caption{
    \textbf{Sensitivity analysis of \textsc{Trace}.}
    \textbf{(a)} A larger attribution frontier initially improves GenEval2
    performance by retaining more plausible responsibilities, but yields
    diminishing gains while increasing the number of candidate edits.
    \textbf{(b)} Increasing $K$ improves attribution reliability by reducing
    the influence of stochastic generation, with gains saturating as generator
    calls increase.
    \textbf{(c)} Increasing $L$ improves held-out performance by enabling
    residual failures to be re-attributed across responsibilities, while deeper
    evolution provides limited additional benefit at higher cost.
    }
    \label{fig:trace_sensitivity}
\end{figure}

As shown in Fig.~\ref{fig:trace_sensitivity}, performance improves rapidly as
the search frontier, trajectory evidence, and evolution depth increase, but the
gains become marginal beyond moderate budgets.
We therefore use $r=3$, $K=4$, and $L=3$ as the default configuration, which
provides a favorable balance between evolution quality and computational cost.

\subsection{Robustness to Different Proposal LLMs}
\label{app:llm_robustness}

We further examine whether EVOGEN-HARNESS depends on a
particular proposal LLM. We replace only the
responsibility-conditioned proposal operator while keeping the Harness Localizer, image generator, verifier,
candidate budget, stochastic seeds, and validation protocol unchanged. We compare the default GPT-4.1 with the
open-weight Qwen3.5-27B and the stronger GPT-5.5.

\begin{table}[t]
\centering
\caption{
\textbf{Robustness to different LLMs on GenEval2.}
Only the proposal LLM is replaced while all other components
and budgets remain fixed.
}
\label{tab:llm_robustness}

\small
\setlength{\tabcolsep}{4.5pt}
\renewcommand{\arraystretch}{1.10}

\begin{tabular}{lcccc}
\toprule
Proposal LLM
& GenEval2 $\uparrow$
& Held-out Gain $\uparrow$
& Commit Rate $\uparrow$
& Regression $\downarrow$ \\
\midrule

Qwen3.5-27B
& 0.6996
& +0.133
& 70.6\%
& 2.3\% \\

GPT-5.5
& \textbf{0.7164}
& \textbf{+0.146}
& \textbf{73.2\%}
& \textbf{1.7\%} \\

\rowcolor{oursbg}
GPT-4.1 (Default)
& 0.7089
& +0.140
& 72.0\%
& 1.9\% \\

\bottomrule
\end{tabular}
\end{table}

Table~\ref{tab:llm_robustness} shows that EVOGEN-HARNESS
remains effective across different proposal LLMs. Qwen3.5-27B
retains performance close to the default GPT-4.1 setting,
while GPT-5.5 provides a modest further improvement.
These results indicate that the effectiveness of
EVOGEN-HARNESS primarily comes from \textsc{TRACE} and its
attribution-guided, regression-safe evolution procedure,
rather than from a particular proposal LLM.

\subsection{Additional Responsibility-Specific Evolution Examples}
\label{app:responsibility_examples}

\begin{figure}[!htbp]
    \centering

    \includegraphics[width=\textwidth]
    {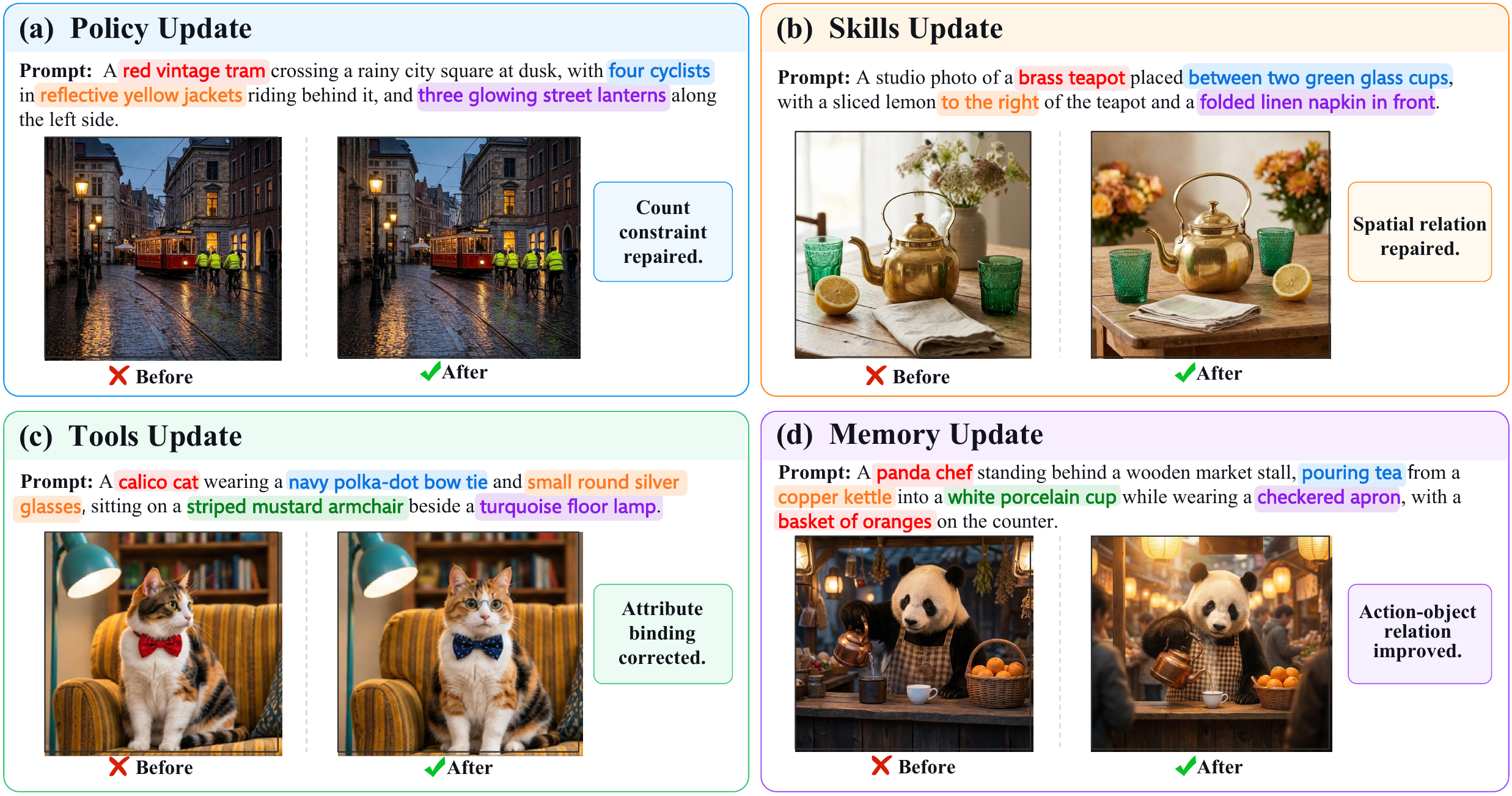}

    \caption{
    \textbf{Responsibility-specific harness evolution.}
    Representative Policy, Skills, Tools, and Memory updates repair distinct
    persistent failure modes while the image generator remains frozen.}
    \label{fig:evolution_examples}
\end{figure}

Figure~\ref{fig:evolution_examples} complements the aggregate
responsibility statistics in Table~\ref{tab:responsibility_analysis} with representative evolution cases.
The examples illustrate how different persistent failure sources induce
different localized harness updates rather than a uniform correction.

Specifically, the Policy example strengthens an explicit counting constraint,
the Skills example revises a reusable procedure for spatial composition,
the Tools example corrects capability knowledge associated with attribute
binding, and the Memory example updates reusable experience for an
action--object relation. In all cases, the underlying image generator remains
frozen; only the responsibility selected by \textsc{Trace} is modified.
Middleware updates are characterized quantitatively in Table~\ref{tab:responsibility_analysis} and illustrated
through the concrete edit example in Appendix~\ref{app:implementation}.

\clearpage
\subsection{Additional Cross-Generator Examples}
\label{app:cross_generator_examples}

\begin{figure}[!htbp]
    \centering

    \includegraphics[width=\textwidth]
    {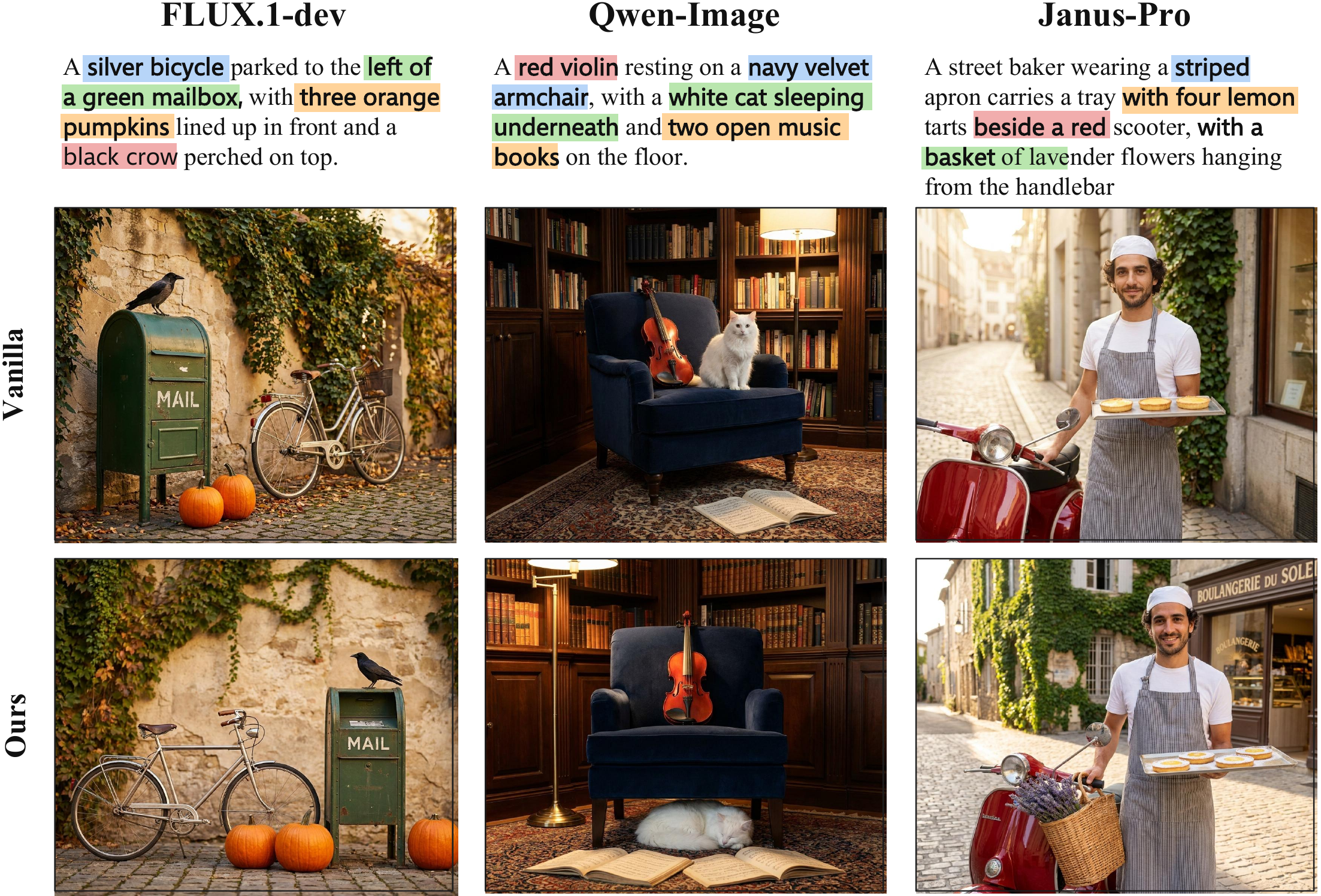}

    \caption{
    \textbf{Cross-generator adaptation.}
    Representative qualitative results of \textsc{EVOGEN-HARNESS} on
    FLUX.1-dev, Qwen-Image, and Janus-Pro. Each image generator remains frozen
    while its generator-external harness is evolved.
    }
    \label{fig:cross_generator_examples}
\end{figure}

Table~\ref{tab:backbone_generalization} quantitatively shows that \textsc{EVOGEN-HARNESS} consistently
improves three frozen T2I backbones. Figure~\ref{fig:cross_generator_examples}
provides representative qualitative examples that complement these results.

For each backbone, we apply the same \textsc{EVOGEN-HARNESS} framework while
keeping the image generator fixed and evolving only its external harness.
The evolved harness is therefore adapted separately to each frozen generator;
we do not assume zero-shot transfer of a single evolved harness across
backbones. The examples span counting, spatial placement, and multi-object
relations, showing that the benefit of harness evolution is not tied to the
rendering behavior of a particular T2I model.

\clearpage
\section{Harness Localizer Prompt}
\label{app:localizer_prompt}

The Harness Localizer in \textsc{Trace} is implemented as a fixed
LLM-based attribution module. It receives trajectory-relative evidence
and the sanitized current harness state, and outputs responsibility-level
scores over five editable harness responsibilities and a \textsc{NO-PATCH}
option. The complete system prompt is provided below.

\begin{tcolorbox}[
    title={Harness Localizer Prompt},
    colback=ReportPale,
    colframe=ReportAccent,
    colbacktitle=ReportPale,
    coltitle=ReportInk,
    fonttitle=\sffamily\bfseries,
    boxrule=0.45pt,
    arc=2pt,
    left=10pt, right=10pt, top=8pt, bottom=8pt,
    breakable,
    enhanced
]
\fontsize{9.6}{11.5}\selectfont
\setlength{\parskip}{3.5pt}

You are the Harness Localizer in \textsc{Trace}, a fixed LLM role in an
attribution-guided evolution system for visual generation agents.

Your task is to identify where a persistent model-external harness
weakness should be investigated. You perform attribution only: visual
verification determines what failed, while you determine which harness
responsibilities could explain the failure.

You do not propose edits, select candidates, validate patches, modify the
generator, or claim that a repair will succeed.

The harness contains five independent responsibilities:

\textbf{1. POLICY}
Policy defines how user requests are interpreted into visual requirements,
constraints, evidence rules, and preservation requirements. Assign higher
scores when failures indicate incorrect requirement representation or
constraint interpretation.

\textbf{2. TOOLS}
Tools describe available generation and verification capabilities,
limitations, and usage conditions. Assign higher scores when evidence
suggests incorrect capability knowledge or missing tool information.

\textbf{3. SKILLS}
Skills represent reusable generation procedures, including decomposition,
prompt construction, layout planning, object binding, and verification-aware
procedures. Assign higher scores when reusable procedures are insufficient
for similar requests.

\textbf{4. MIDDLEWARE}
Middleware controls runtime orchestration, including routing, skill
selection, retries, verification, and execution ordering. Assign higher
scores when available resources are sufficient but the execution strategy
is inappropriate.

\textbf{5. MEMORY}
Memory stores reusable cross-task experience, recurring failure patterns,
and validated corrective knowledge. Assign higher scores when failures
indicate missing transferable experience.

\textbf{6. NO-PATCH}
\textsc{NO-PATCH} indicates that persistent evolution is not justified.
Assign higher scores when failures are better explained by stochastic
generation, ambiguous intent, verifier uncertainty, or insufficient
evidence.

Perform attribution following these principles:

(1) Identify the violated visual constraint rather than only the observed
symptom.

(2) Consider repeated evidence across stochastic executions.

(3) Attribute failures to the earliest responsibility-specific decision
that could plausibly explain the failure.

(4) Distinguish requirement interpretation, capability knowledge, reusable
procedures, runtime orchestration, and persistent experience.

(5) Preserve multiple plausible causes when evidence supports interacting
responsibilities.

(6) Prefer \textsc{NO-PATCH} when persistent evolution is not supported.

Return independent scores between 0 and 1 for all responsibilities.
The scores represent evidence strength rather than probabilities and do
not need to sum to one.

Return only one valid JSON object:

\begin{verbatim}
{
  "policy": score,
  "tools": score,
  "skills": score,
  "middleware": score,
  "memory": score,
  "no_patch": score,
  "rationale": "brief evidence-based explanation"
}
\end{verbatim}

The rationale should explain the strongest responsibility, mention
uncertainty when applicable, and justify whether \textsc{NO-PATCH} is
preferred.

\end{tcolorbox}
\end{document}